\documentclass{article}

\usepackage{multicol}

\makeatletter
\let\NAT@parse\undefined
\makeatother

\usepackage{comment}
	\usepackage{amsmath}
	\usepackage{verbatim} 
	\usepackage{amsfonts}
	\usepackage{amssymb}
	\usepackage{ragged2e}
	\usepackage{graphicx}
	\usepackage{cancel}
	\usepackage{mathtools}
	\usepackage{tabularx}
     \usepackage{adjustbox} 
	\usepackage{arydshln}
	\usepackage{tensor}
	\usepackage{array}
	\usepackage[dvipsnames]{xcolor}
	\usepackage{listings}
	\usepackage{textcomp}
	\usepackage{mathrsfs}
	\usepackage{bbm}
	\usepackage{tikz}
	\usepackage{tikz-cd}
	\usepackage{enumitem}
	\usepackage{arydshln}
	\usepackage{relsize}
	\usepackage{multirow}
	\usepackage{scalerel}
	\usepackage{upgreek}
	\usepackage{ifthen}
	\usepackage{yhmath}
	\usepackage{blkarray}
	\usepackage{dashrule}
	\usepackage{subcaption}
	\usepackage{overpic}
    \usepackage[commandnameprefix=ifneeded]{changes}

	\usepackage{letltxmacro}
	\LetLtxMacro\orgvdots\vdots
	\LetLtxMacro\orgddots\ddots

	\makeatletter
	\DeclareRobustCommand\vdots{%
		\mathpalette\@vdots{}%
	}
	\newcommand*{\@vdots}[2]{%
		\sbox0{$#1\cdotp\cdotp\cdotp\m@th$}%
		\sbox2{$#1.\m@th$}%
		\vbox{%
			\dimen@=\wd0 %
			\advance\dimen@ -3\ht2 %
			\kern.5\dimen@
			\dimen@=\wd2 %
			\advance\dimen@ -\ht2 %
			\dimen2=\wd0 %
			\advance\dimen2 -\dimen@
			\vbox to \dimen2{%
				\offinterlineskip
				\copy2 \vfill\copy2 \vfill\copy2 %
			}%
		}%
	}
	\DeclareRobustCommand\ddots{%
		\mathinner{%
			\mathpalette\@ddots{}%
			\mkern\thinmuskip
		}%
	}
	\newcommand*{\@ddots}[2]{%
		\sbox0{$#1\cdotp\cdotp\cdotp\m@th$}%
		\sbox2{$#1.\m@th$}%
		\vbox{%
			\dimen@=\wd0 %
			\advance\dimen@ -3\ht2 %
			\kern.5\dimen@
			\dimen@=\wd2 %
			\advance\dimen@ -\ht2 %
			\dimen2=\wd0 %
			\advance\dimen2 -\dimen@
			\vbox to \dimen2{%
				\offinterlineskip
				\hbox{$#1\mathpunct{.}\m@th$}%
				\vfill
				\hbox{$#1\mathpunct{\kern\wd2}\mathpunct{.}\m@th$}%
				\vfill
				\hbox{$#1\mathpunct{\kern\wd2}\mathpunct{\kern\wd2}\mathpunct{.}\m@th$}%
			}%
		}%
	}
	\makeatother

	\tikzset{
	  symbol/.style={
		draw=none,
		every to/.append style={
		  edge node={node [sloped, allow upside down, auto=false]{$#1$}}}
	  }
	}

\usepackage[bookmarks=true]{hyperref}

\usepackage{cleveref}
\Crefname{figure}{Figure}{Figures}
\Crefname{table}{Table}{Tables}
\Crefname{equation}{Eq.}{Eqs.}
\Crefname{section}{Section}{Sections}
\Crefname{subsection}{Subsection}{Subsections}
\Crefname{appendix}{Appendix}{Appendices}

\usepackage[commentColor=black]{algpseudocodex}
\tikzset{algpxIndentLine/.style={draw=black}}
\algrenewcommand{\alglinenumber}[1]{\bfseries\footnotesize #1}
\algrenewcommand{\textproc}{}
\algrenewcommand{\algorithmicrequire}{\textbf{Input:}}
\algrenewcommand{\algorithmicensure}{\textbf{Output:}}
\usepackage{algorithm}
\makeatletter
\newcommand{\algorithmname}{\ALG@name}
\renewcommand{\floatc@ruled}[2]{{\@fs@cfont #1:} #2\par}
\makeatother

\usetikzlibrary{shapes.geometric}
\usetikzlibrary{angles, quotes, calc}
\usetikzlibrary{arrows.meta}

\definecolor{awardcolor}{HTML}{f5586a}
\definecolor{darkbeautifulblue}{RGB}{65, 105, 225} 

\usepackage{amssymb}
\usepackage{times}

\usepackage{multicol}
\usepackage[bookmarks=true]{hyperref}
\usepackage{amsmath}
\usepackage{booktabs}
\usepackage{multirow}
\usepackage{graphicx}
\usepackage{xspace}
\usepackage{xcolor}
\usepackage{caption}
\usepackage{wrapfig}
\usepackage{bbding}  
\usepackage{pifont}  
\usepackage{float}
\usepackage{amssymb}  

\newcommand{\CustomDash}{\textcolor{gred}{\rule[0.5ex]{0.8em}{0.23em}}} 

\newcommand{\oursfull}[0]{Tool-as-Interface\xspace}

\usepackage{colortbl}
\definecolor{ourcolor}{HTML}{99e0eb}
\definecolor{ourblue}{HTML}{27a2c3}

\definecolor{tablecolor}{HTML}{ccf2f5} 

\definecolor{tablecolor2}{HTML}{ffcdb4}
\definecolor{citecolor}{HTML}{fe7b5b}
\definecolor{grey}{rgb}{0.9, 0.9, 0.9}

\usepackage{listings}
\definecolor{gred}{rgb}{0.859,0.267,0.216}
\definecolor{ggreen}{rgb}{0.059,0.616,0.345}

\definecolor{deepblue}{HTML}{27a2c3}

\definecolor{deepred}{HTML}{fe7b5b}

\usepackage{tikz}

\definecolor{ElectricBlue}{HTML}{007BFF}
\definecolor{BrightOrange}{HTML}{FFA500}
\definecolor{LimeGreen}{HTML}{32CD32}
\definecolor{CrimsonRed}{HTML}{DC143C}

\usepackage[preprint]{corl_2025} 

\title{Tool-as-Interface: Learning Robot Policies from Observing Human Tool Use}

\author{
Haonan Chen$^1$, Cheng Zhu$^1$, Shuijing Liu$^2$, Yunzhu Li$^3$, Katherine Driggs-Campbell$^1$ \\ 
$^1$ University of Illinois, Urbana-Champaign \quad $^2$ UT Austin \quad 
$^3$ Columbia University 
\\ \href{https://tool-as-interface.github.io}{https://tool-as-interface.github.io}
}

\begin{document}
{%
\renewcommand\twocolumn[1][]{#1}%
\maketitle
\vspace{-0.08cm}
\begin{center}
    \centering
    \captionsetup{type=figure}
     \includegraphics[width=0.95\textwidth]{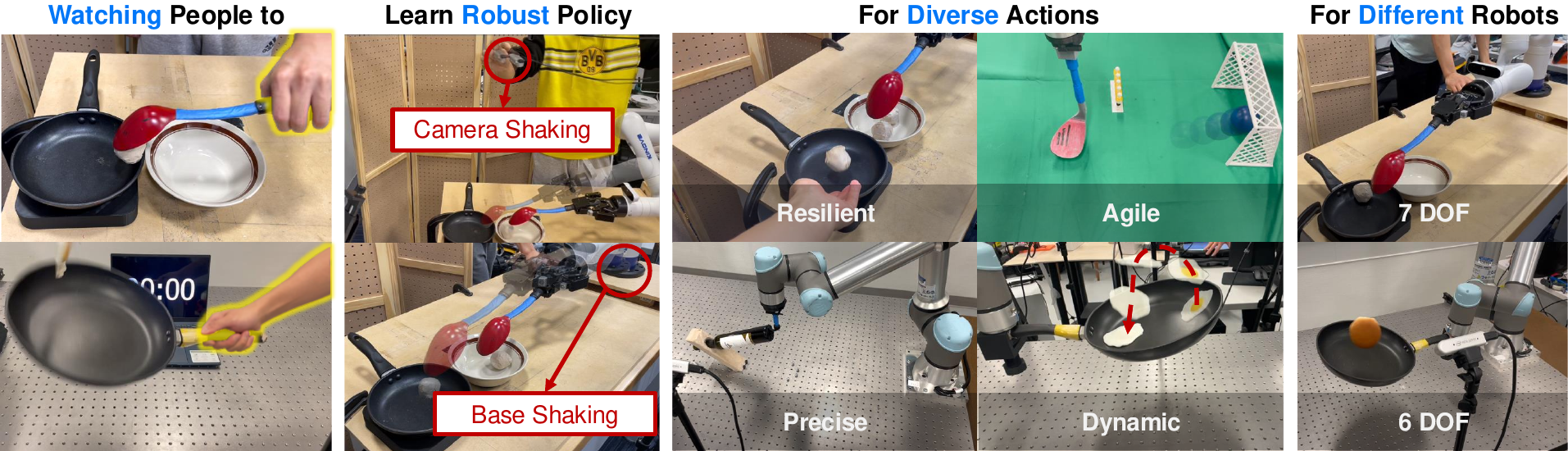}
     \vspace{-0.05in}
    \caption{\scriptsize{\textbf{\oursfull.} We propose a scalable data collection and policy learning framework designed to transfer diverse, intuitive, and natural human data into effective visuomotor policies. The framework enables robots to learn robust policies that can operate effectively under challenging conditions, such as base and camera movement, and achieve high performance on a variety of complex manipulation tasks.}}
    \label{fig:teaser}
\end{center}
\vspace{-0.1in}
}

\begin{abstract}

    Tool use is essential for enabling robots to perform complex real-world tasks, but learning such skills requires extensive datasets.
    While teleoperation is widely used, it is slow, delay-sensitive, and poorly suited for dynamic tasks. In contrast, human videos provide a natural way for data collection without specialized hardware, though they pose challenges on robot learning due to viewpoint variations and embodiment gaps.
    To address these challenges, we propose a framework that transfers tool-use knowledge from humans to robots. To improve the policy's robustness to viewpoint variations, we use two RGB cameras to reconstruct 3D scenes and apply Gaussian splatting for novel view synthesis. 
    We reduce the embodiment gap using segmented observations and tool-centric, task-space actions to achieve embodiment-invariant visuomotor policy learning.
    We demonstrate our framework's effectiveness across a diverse suite of tool-use tasks, where our learned policy shows strong generalization and robustness to human perturbations, camera motion, and robot base movement. Our method achieves a 71\% improvement in task success over teleoperation-based diffusion policies and dramatically reduces data collection time by 77\% and 41\% compared to teleoperation and the state-of-the-art interface, respectively.




\end{abstract}


\section{Introduction} 

Tool use enables humans to perform complex tasks by extending their physical capabilities. In contrast, robotic systems remain largely limited to grasping and pick-and-place operations~\cite{Vaesen_2012, OsiurakHeinke2018, 844081, Prattichizzo2016, carbone2012grasping, 16222, robotics10030105}. To enable richer manipulation skills, robots must learn to use diverse tools in dynamic environments. This work focuses on the efficient training of robot policies for tool use, with an emphasis on scalable and low-cost data collection. 

Imitation learning (IL) is a promising way to acquire tool-use skills from human demonstrations~\cite{Fang2019, s21041278, jang2022bc}. Prior work has leveraged teleoperation platforms~\cite{wang2023mimicplay, brohan2022rt, jang2022bc} and hand-held grippers~\cite{seo2024legato, chi2024universal} to provide precise supervision. However, these systems often require expensive hardware, 3D-printed tools, or expert calibration, hindering scalability across diverse users and environments.



As a scalable alternative, we learn from natural human manipulation videos. While abundant, this data introduces challenges from viewpoint differences and human-robot embodiment gaps~\cite{shao2021concept2robot, bahl2022human, pmlr-v205-pan23a}. 
We introduce a new framework that leverages two-view human manipulation videos and appropriate representations of states and actions  to train robot policies for tool use. To close the viewpoint gap, our system uses 3D scene reconstruction and novel view synthesis to train a viewpoint-invariant policy. To bridge the embodiment mismatch, we filter out embodiment-specific features like hands using segmentation, and employ a task-space, tool-centric action representation supports robustness to robot base variation (Figure~\ref{fig:teaser}).

Our contributions are as follows: (1) We introduce a framework for scalable, intuitive, and cost-effective data collection for robot tool-use learning, using two-view human manipulation videos without requiring teleoperation or specialized hardware; (2) We demonstrate strong generalization across diverse real-world tool-use tasks (e.g., nail hammering, meatball scooping, pan flipping, wine bottle balancing, and soccer ball kicking) achieving a 71\% higher success rate and 77\% reduction in data collection time compared to diffusion policies trained on SpaceMouse~\cite{3dconnecxion2023spacemouse} or Gello~\cite{wu2023gello}, and a 41\% improvement over handheld grippers like UMI~\cite{chi2024universal}; and (3) We provide a detailed robustness analysis, evaluating performance under changes in viewpoint, robot base configuration, and human motion, along with ablations on segmentation, novel view synthesis, and random cropping.

\section{Related Works}

\noindent\textbf{Data Collection for Robot Learning:}
High-quality data is essential for teaching robots new skills. Simulation offers scalability and low cost~\cite{Zhao2020SimtoRealTI, 8460528, doi:10.1177/0278364920987859, 8968053}, but the sim-to-real gap remains challenging. Learning from real-world teleoperated demonstrations helps train policies with minimal distribution shift between training and testing~\cite{zhu2022viola, chi2023diffusionpolicy, chi2024diffusionpolicy, seo2023trill, Wong2021ErrorAwareIL}. Leader-follower systems (e.g., ALOHA~\cite{zhao2023learning, fu2024mobile}, GELLO~\cite{wu2023gello}) simplify teleoperation by providing intuitive kinematic replication but require real robots and are costly. Portable hand-held grippers (e.g., UMI~\cite{chi2024universal}, LEGATO~\cite{seo2024legato}) enable flexible data collection but still require specialized hardware. Tool-based policy representations offer another direction. MimicTouch leverages tactile feedback for contact-rich skills~\cite{yu2024mimictouch}, while ScrewMimic models bimanual tasks as constrained screw motions for learning from human videos~\cite{bahety2024screwmimic}. However, tactile methods require extra hardware, and the screw motion assumption may not hold. Another approach transfers single-demonstration trajectories across objects but assumes static object configurations~\cite{wen2022yodo}, which will not work with changing spatial configurations. In contrast, our method uses natural human demonstrations without tactile sensors, special tools, or motion constraints—enabling scalable, low-cost data collection in unconstrained environments. \looseness=-1

\noindent\textbf{Cross-Embodiment Policy Learning:}
Cross-embodiment learning enables policy transfer across robots with diverse kinematic structures~\cite{hu2022know, salhotra2023bridging, yang2023polybot}. Prior approaches relying on multi-embodiment datasets~\cite{devin2016learning, wang2016nervenet, yu2017preparing, chen2019hardware, ghadirzadeh2021bayesian, salhotra2023bridging, attarian2023geometry} focus solely on data from different robot embodiments and struggle to leverage human demonstrations effectively. Recent methods tokenize observations and actions into unified transformer networks~\cite{sferrazza2024body, yang2024pushinglimitscrossembodimentlearning}, but require large models, extensive datasets, and cannot directly transfer policies across embodiments.
Other efforts use human data to estimate point flow~\cite{wen2023anypoint} or generate high-level plans~\cite{lynch2020learning, wang2023mimicplay}, but still depend on robot-generated data for low-level control. Additionally, prior works emphasize visual consistency through embodiment masking~\cite{bahl2022human, kareer2024egomimic}. However,  \citet{bahl2022human} relies on predefined motion primitives and \citet{kareer2024egomimic} requires robot data for augmentation. As a step further, our approach adopts a similar masking idea but enables robots to learn freely, even agile motions from human videos without any robot data. \looseness=-1

\noindent\textbf{Cross-Viewpoint Policy Learning:}
Robots can encounter viewpoint variations when interacting with their environments. Prior work addresses this by learning view-invariant representations~\cite{sermanet2018time, chen2021unsupervised}, equivariant 3D features~\cite{shridhar2023perceiver, gervet2023act3d, zhu2023learning}, or augmenting datasets with varied viewpoints~\cite{sadeghi2018sim2realservo, zhou2023nerf}. 
An early attempt by \citet{NEURIPS2019_8a146f1a} introduced a hierarchical model mapping third-person demonstrations to first-person sub-goals, with a low-level module predicting actions.
Recent work by \citet{chen2024roviaug} leverages SAM~\cite{kirillov2023segany} and ControlNet~\cite{zhang2023adding} to transform robot images across different robots, employing ZeroNVS~\cite{zeronvs} for novel-view synthesis to augment datasets and improve policy generalization across viewpoints.
Another approach by \citet{yuan2024learning} integrates reinforcement learning, multi-view representation learning, and a Spatial Transformer Network to enhance policy robustness in visually complex environments. Despite promising results, these methods face scalability issues, reliance on computationally intensive view synthesis, or the need for tailored simulators. In contrast, our approach achieves efficient and scalable cross-viewpoint transfer directly from human data by leveraging the MASt3R~\cite{mast3r_arxiv24} foundation model for 3D reconstruction from two RGB images and employing Gaussian splatting for fast, cost-effective novel-view synthesis.\looseness=-1


\section{Tool-as-Interface Framework}
\label{sec:method}

\begin{figure*}[tb]
    \centering
    \includegraphics[width=1\textwidth]{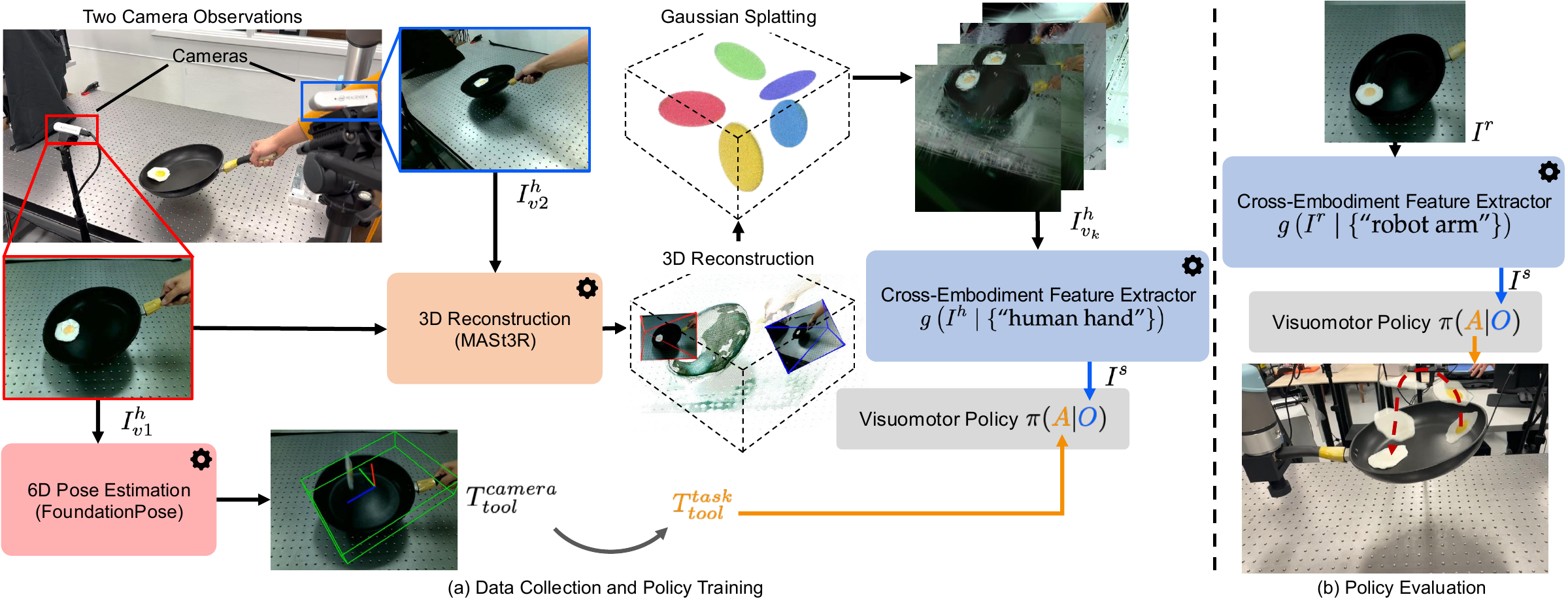}
    \vspace{-0.23in}
    \caption{\scriptsize{\textbf{Policy Design.} Human manipulation data was collected using two cameras and processed through the foundation model MASt3R~\cite{mast3r_arxiv24} to generate 3D reconstructions. Using 3D Gaussian splatting, we sampled novel views to augment the dataset. Human hands were segmented to create embodiment-agnostic observations as policy inputs. For action labeling, FoundationPose~\cite{foundationposewen2024} estimated the tool’s pose in the camera frame, $T_\text{tool}^\text{camera}$, which was transformed into task space, $T_\text{tool}^\text{task}$. A diffusion model was then trained as the visuomotor policy.}}
    
    \label{fig:method}
    \vspace{-0.25in}
\end{figure*}


\noindent\textbf{Problem Statement:}
We formulate robotic manipulation as a Markov Decision Process (MDP), where the goal is to learn a policy $\pi: \mathcal{O}^r \to \mathcal{A}$ that enables a robot to perform a given task. 
The robot’s observation space $\mathcal{O}^r$ consists of a single-view RGB image $I^r \in \mathcal{I}^r$ and proprioceptive data $x^r \in SE(3)$, where each $I^r$ is a tensor in $\mathbb{R}^{128 \times 128 \times 3}$.
We train the policy using an imitation dataset of $N$ human demonstrations, $D = {(\mathcal{O}_0^h, \mathcal{O}_1^h, \dots)}_{n=1}^N$, where each $\mathcal{O}^h = \{I_{v1}^h, I_{v2}^h\}$ contains two RGB images captured from different viewpoints and each $I_{vi}^h\in\mathcal{I}^h$ is a tensor in $\mathbb{R}^{480 \times 640 \times 3}$.
We preprocess the dataset to infer actions using a 6D pose estimation and tracking model, resulting in $D = \{(\mathcal{O}_0^h, a_0, \mathcal{O}_1^h, a_1, \dots)\}_{n=1}^{N}$, where each action $a \in SE(3)$.
To bridge the embodiment gap between humans and robots, we assume the tool is rigidly attached to both the human hand (implicitly) and the robot end-effector (explicitly), with a fixed transformation estimated prior to deployment. Under this setup, the robot can reproduce human-demonstrated tool trajectories, enabling policy transfer across embodiments while preserving task-relevant behaviors (Figure~\ref{fig:method}).\looseness=-1

\noindent\textbf{Tool-Centric Demonstrations for Robot Manipulation:}
We leverage the fact that both humans and robots can operate the same physical tools to facilitate policy learning. Tools serve as a shared interface for interacting with objects, enabling the direct transfer of human demonstrations with minimal embodiment-specific adaptation. Unlike prior work focused on grasping or pick-and-place tasks~\cite{chi2024universal,seo2024legato,young2020visual}, our approach enables robots to perform complex interactions using everyday tools. Our formulation abstracts actions to the tool pose, reducing morphological dependence and promoting policy generalization across embodiments. It also simplifies data collection by eliminating the need for robot-specific demonstrations. Humans can naturally manipulate tools by hand without extra instrumentation. For deployment, robots either rigidly grasp the tool, as demonstrated with a Kinova Gen3 arm, or attach it using a custom fast tool changer described in Appendix~\ref{appendix:hardware_design} and shown in Figure~\ref{fig:tool_changer}, compatible with ISO 9409-1-50-4-M6 flanges.

\noindent\textbf{Perception Alignment Across Embodiments:}
To enable cross-embodiment policy transfer, we align human and robot observations within a shared visual space $\mathcal{I}^s$ by applying a feature extraction function $g: \mathcal{I}^h \cup \mathcal{I}^r \to \mathcal{I}^s$. We instantiate $g$ with Grounded-SAM~\cite{ren2024grounded}, using prompts such as “human hand” and “robot arm” to mask out embodiment-specific regions—human hands during training and robotic arms during deployment. Masking out these regions ensures that only task-relevant visual information (e.g., tools and objects) remains visible in both phases. By minimizing visual discrepancies between human and robot data, the feature extraction process reduces embodiment-specific bias and improves generalization across embodiments.


\noindent\textbf{3D-Aware View Augmentation:}
We use cameras for data collection due to their widespread availability—over 7.14 billion smartphones are equipped with them~\cite{Jonsson2024}. However, single-camera setups suffer from scale ambiguity and limited 3D perception and are sensitive to viewpoint changes.


\noindent\textsc{3D Reconstruction:} To address this, we use MASt3R~\cite{mast3r_arxiv24}, an image-matching model that reconstructs accurate 3D environments from just two RGB images—eliminating the need for depth sensors, which are less common and more power-hungry. Two cameras suffice to avoid scale ambiguity inherent in monocular settings. MASt3R produces high-quality point clouds without requiring known camera extrinsics or intrinsics by globally aligning multi-view features.



\noindent\textsc{View Synthesis and Augmentation:} 
3D Gaussian splatting synthesizes novel viewpoints from the reconstructed scene, allowing the robot to observe interactions from multiple angles—even when only two views are available. The resulting perspectives augment the training data, increasing visual diversity and improving policy learning. To further enhance robustness and generalization, random cropping is applied, following diffusion policy~\cite{chi2023diffusionpolicy, chi2024diffusionpolicy}.

\begin{wrapfigure}{r}{0.33\textwidth} 
\vspace{-1.8em}
\centering
\includegraphics[width=0.33\textwidth]{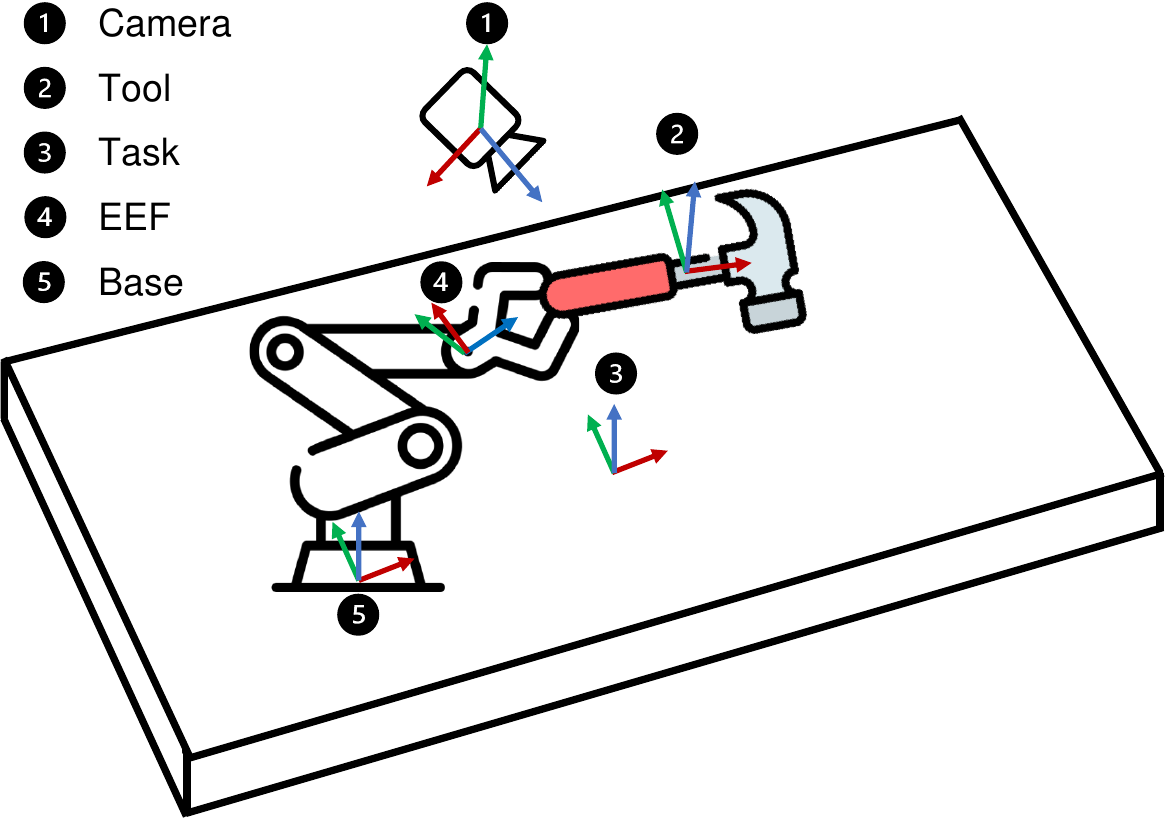}
\vspace{-1.5em}
\caption{\scriptsize\textbf{Coordinate System Diagram.} The diagram shows the Camera, Tool, Task space, End-Effector (EEF), and Base frames. The action is represented as \(T_{\text{tool}}^{\text{task}}\).}
\label{fig:coordinate_system_diagram}
\vspace{-1em}
\end{wrapfigure}

\noindent\textbf{Tool-Centric Action Representation and Policy Deployment:}  
To support general tool usage, we propose a task-frame, tool-centric action representation denoted as $T_{\text{tool}}^{\text{task}}$, which describes the tool's motion independently of human or robot morphology, camera pose, or base configuration. This invariant formulation enables robust policy transfer across different embodiments and viewpoints. As shown in Figure~\ref{fig:coordinate_system_diagram}, the tool's pose is first estimated in the camera frame using a 6D pose estimation model (e.g., FoundationPose~\cite{foundationposewen2024}) as $T_{\text{tool}}^{\text{camera}}$, and then transformed into the task frame:
\[
T_{\text{tool}}^{\text{task}} = T_{\text{camera}}^{\text{task}} T_{\text{tool}}^{\text{camera}},
\]
where $T_{\text{camera}}^{\text{task}}$ denotes the transformation from the camera to the task frame.  

A diffusion policy~\cite{chi2023diffusionpolicy} maps a single-view RGB image to a predicted SE(3) action \(T_{\text{tool}}^{\text{task}}\). At deployment, the robot command is computed by converting the prediction to the end-effector frame. For stationary robots, the task frame aligns with the base frame; for mobile platforms, base movement is compensated using \(T_{\text{task}}^{\text{base}}\). The resulting end-effector pose is given by:
\[
T_{\text{eef}}^{\text{base}} = T_{\text{task}}^{\text{base}} T_{\text{tool}}^{\text{task}} T_{\text{eef}}^{\text{tool}},
\]
where $T_{\text{eef}}^{\text{tool}}$ is the known fixed transform between the tool and the robot end-effector.


\section{Policy Evaluations}

Our evaluations aim to assess our framework across three dimensions: \textbf{reliability} (how consistently and successfully the learned policies perform), \textbf{execution efficiency} (how smooth and natural the resulting behaviors are), and \textbf{versatility} (how well the framework adapts to diverse tasks and generalizes across conditions).

\noindent\textbf{Experimental Tasks Overview:} We evaluate five real-world robotic tasks on Kinova Gen3 and UR5e robots, involving precision manipulation, dynamic object handling, and dexterous tool use. Policies use RGB inputs from RealSense D415 cameras and handle variations in object positions and camera poses. Tasks include: (1) {Nail Hammering} – Precise striking of a small target, (2) {Meatball Scooping} – Contact-sensitive rolling object manipulation, (3) {Pan Flipping} – Fast, dynamic flipping with varied objects, (4) {Wine Balancing} – Gravity-aware placement into an unstable rack, and (5) {Soccer Ball Kicking} – Dynamic interception and obstacle avoidance. Full details in Appendix~\ref{appendix:tasks}.

\noindent\textbf{Baselines:}
We evaluate the effectiveness and efficiency of learning directly from human manipulation videos without relying on robot-generated data. We benchmark against a diffusion policy trained on robot demonstrations and UMI~\cite{chi2024universal}, a hand-held gripper method. Robot demonstrations are collected using SpaceMouse or Gello~\cite{wu2023gello} under identical time budgets. Additionally, we conduct ablations to assess random cropping before policy training, novel view synthesis data augmentation, and embodiment segmentation. To further illustrate the advantages of our approach, we compare trajectory rollouts for a meatball-scooping episode, highlighting how our method is more sample-efficient and less prone to distribution shifts by eliminating excessive waypoints.

\noindent\textbf{Evaluation Metrics:}
During testing, we introduce two types of variations: (1) randomizing the initial spatial configurations of objects in each task to assess policy generalization, and (2) varying camera positions to evaluate the robustness of policies to different viewpoints. All methods, including the baseline and ablation variants, are tested under the same conditions. Performance is evaluated using two metrics: success rate, which measures the proportion of successfully completed task trials and reflects policy effectiveness, and task completion time, which captures the average duration to complete tasks and reflects policy efficiency.

\vspace{-0.05in}
\section{Experiment Results}


\noindent\textbf{Capabilities and Effectiveness:} 
Table~\ref{tab:success_rates_times} summarizes our real-world results, showing that our framework consistently achieves higher success rates across all tasks compared to baselines. We also compare against the stronger hand-held gripper baseline UMI~\cite{chi2024universal} (Table~\ref{tab:umi_compare}).
In our default setup, SLAM-based mapping failed due to low environmental texture, so we added a textured background to support reliable mapping for UMI. For the nail hammering task, we evaluated UMI with 25 demonstrations (matching our collection time) and 100 demonstrations (to assess ideal performance). UMI fails all 13 trials with 25 demonstrations but succeeded with 100. 
It was also inapplicable \begin{wraptable}{r}{0.42\textwidth}
\vspace{-1em}
\centering
\caption{\scriptsize\textbf{Task Success Rates and Completion Times.} Success rates show completed trials over total. ``DP'' is trained on teleoperation; ``Not Feasible'' means failure.}
\label{tab:success_rates_times}
\vspace{-0.5em}
\scriptsize
\resizebox{0.40\textwidth}{!}{%
\begin{tabular}{@{\hskip 2pt}l@{\hskip 4pt}c@{\hskip 4pt}c@{\hskip 4pt}c@{\hskip 2pt}}
\toprule
\textbf{Task} & \textbf{Method} & \textbf{Success} & \textbf{Time (s)} \\
\midrule
\multirow{2}{*}{Hammer}       & DP   & 0/13           & --         \\
                              & Ours & \textbf{13/13} & \textbf{11.0} \\
\multirow{2}{*}{Scoop}        & DP   & 5/12           & 42.0       \\
                              & Ours & \textbf{10/12} & \textbf{12.4} \\
\multirow{2}{*}{Pan: Egg}     & DP   & Not Feasible   & --         \\
                              & Ours & \textbf{12/12} & \textbf{1.5}  \\
\multirow{2}{*}{Pan: Bun}     & DP   & Not Feasible   & --         \\
                              & Ours & \textbf{9/12}  & \textbf{1.9}  \\
\multirow{2}{*}{Pan: Patty}   & DP   & Not Feasible   & --         \\
                              & Ours & \textbf{10/12} & \textbf{2.3}  \\
\multirow{2}{*}{Wine Balance} & DP   & Not Feasible   & --         \\
                              & Ours & \textbf{8/10}  & \textbf{30.9} \\
\multirow{2}{*}{Soccer Kick}  & DP   & Not Feasible   & --         \\
                              & Ours & \textbf{6/10}  & \textbf{2.0}  \\
\bottomrule
\end{tabular}
}
\vspace{-2em}
\end{wraptable} to wine balancing and pan flipping due to contact and inertial challenges, and struggled 
in soccer kicking due to localization failures. In contrast, our method demonstrates reliable performance across all tasks: accurately detecting spatial locations (nail hammering, meatball scooping), performing high-speed motions (pan flipping), precisely inserting wine bottles, and swiftly reacting in soccer kicking. This strong performance is enabled by collecting significantly larger and more diverse episodes within the same data collection timeframe (see Section~\ref{data_collection}), enabling robust policy training. Our approach overcomes limitations of teleoperation tools like Gello and SpaceMouse, enabling data collection for scenarios they struggle to handle. Qualitative policy rollouts are shown in Figure~\ref{fig:policy_rollout} in Appendix.

\noindent\textbf{Policy Execution Efficiency:} 
Our framework demonstrates high execution efficiency, achieving 
shorter task completion times and smoother action trajectories than baselines (Table~\ref{tab:success_rates_times}).
This efficiency stems from the natural human data, which captures the fluidity and speed of real-world activities, resulting in higher-quality training trajectories. 
In contrast, teleoperated demonstrations \begin{wraptable}{r}{0.44\textwidth}
\vspace{-0.8em}
\centering
\caption{\scriptsize\textbf{Task success rates comparing our method with the hand-held gripper-based method on Nail Hammering.}}
\label{tab:umi_compare}
\vspace{-0.5em}
\scriptsize
\begin{tabular}{@{}lcc@{}}
\toprule
\textbf{Method} & \textbf{Demo Time \& Count} & \textbf{Success} \\
\midrule
UMI~\cite{chi2024universal} & 180s (25) & 0/13 \\
UMI                         & 720s (100) & \textbf{13/13} \\
Ours                        & 180s (40)  & \textbf{13/13} \\
\bottomrule
\end{tabular}
\vspace{+0.4em}
\end{wraptable}
    
\vspace{-1.7em}often produce slower, less natural motions, limiting dynamic performance.
By leveraging more realistic data, our framework accelerates execution while enhancing motion quality, making it suited for real-world applications. See Appendix~\ref{appendix:execution_smooth} for a detailed comparison of policy rollouts.



\noindent\textbf{Benefits of Tool-Based Action Representation in Task Space:} 
Using the tool pose in the camera frame works with a static camera but fails under camera movement due to unreliable real-time tracking and incorrect end-effector positioning in the base frame. Similarly, representing actions in the base frame fails under base movement due to the assumption of a fixed base-to-workspace transform. In contrast, representing actions in task space is invariant to both camera and base movement, enabling robust execution even under large viewpoint shifts and base movements. \looseness=-1

\begin{figure*}[t]  
    \centering
    \includegraphics[width=\textwidth, trim=0pt 0pt 0pt 0pt, clip]{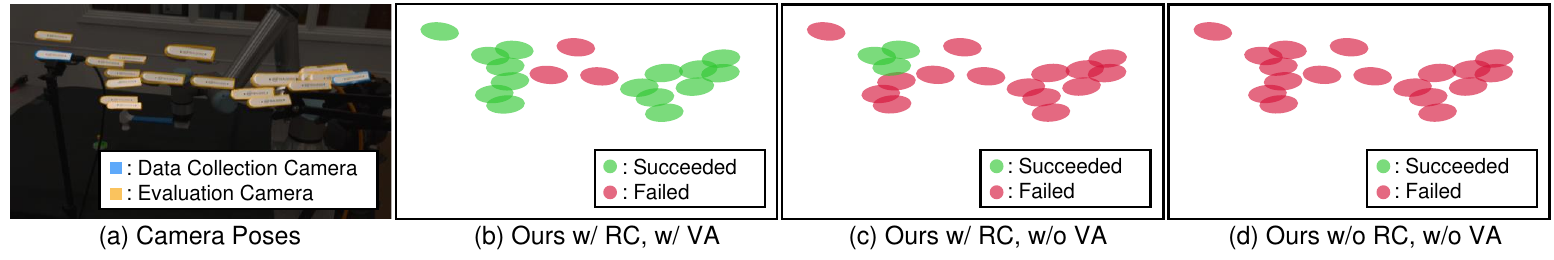}
    \vspace{-0.25in}
    \caption{\scriptsize{\textbf{Policy Testing Across Camera Poses in Nail Hammering.} (a) Camera poses for data collection and evaluation. (b-d) Performance ranges for methods trained with/without random cropping (RC) and view augmentation (VA).}}
    \label{fig:camera_overlay}
    \vspace{-0.148in}
\end{figure*}

\begin{figure*}[t]
    \centering
    \includegraphics[width=1\textwidth]{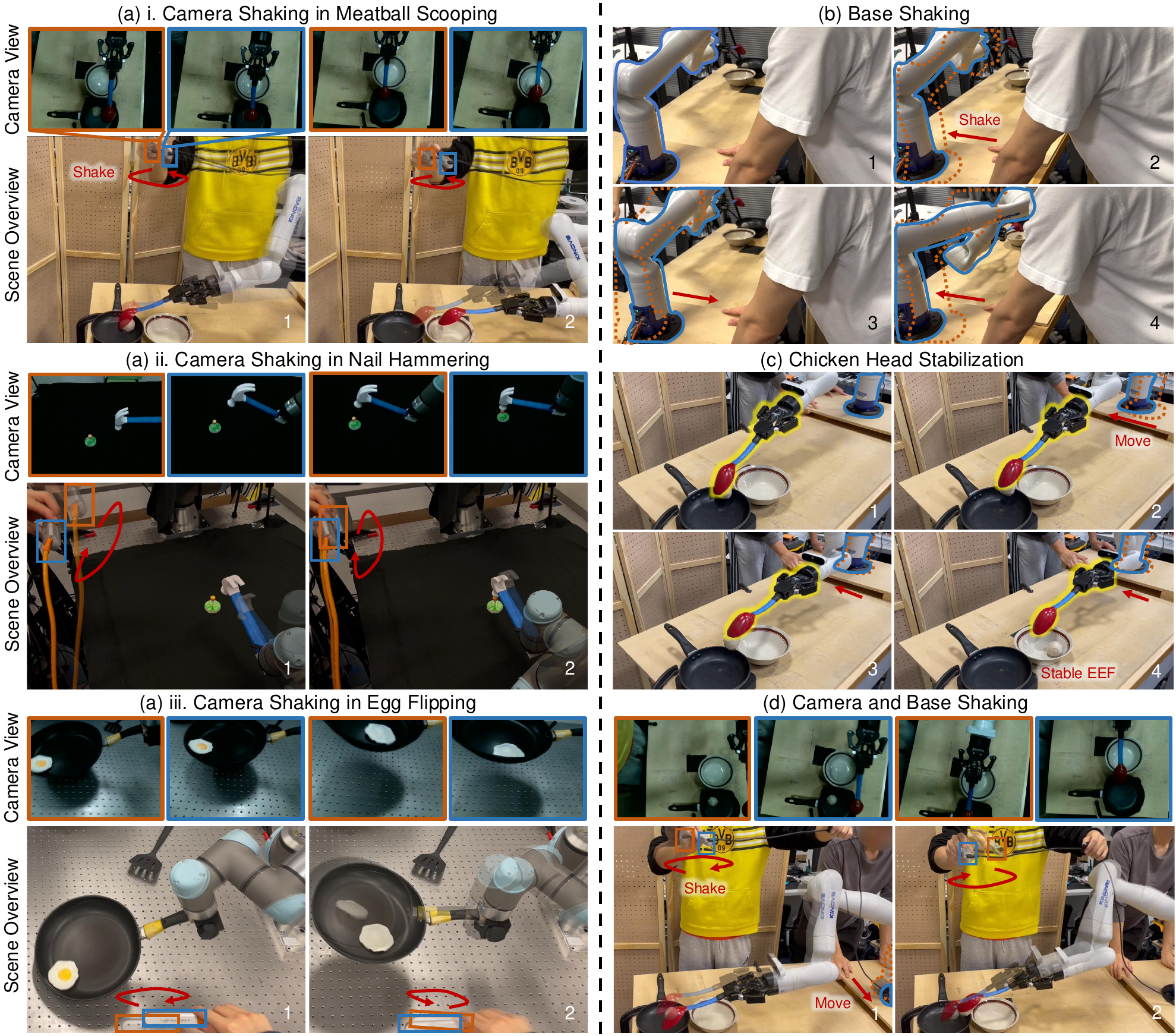}
    \vspace{-0.22in}
    \caption{\scriptsize{\textbf{Robustness to Camera and Base Movement.} (a) Camera Pose Robustness: The policy demonstrated the ability to handle camera shaking across three tasks—meatball scooping, nail hammering, and pan flipping. The first row shows the camera view, while the second row provides a scene overview with the shaking motion. (b) Robot Base Robustness: The policy successfully compensated for base shaking, even when the shaking frequency exceeded the robot's control frequency. (c) Chicken Head Stabilization: At lower base movement frequencies, the end effector displayed a stabilization effect similar to a chicken’s steady head. (d) Combined Robustness: The policy maintained task performance under simultaneous camera and base shaking.}}
    \label{fig:robustness}
    \vspace{-0.26in}
\end{figure*}

\noindent\textbf{Effects of Random Cropping and View Augmentation:} 
Our experiments show that random cropping (RC) and view augmentation (VA) together enhance policy robustness to camera pose variations.
RC improves resilience to minor perturbations such as small movements or shaking, while VA exposes the model to a broader distribution of viewpoints during training.
We evaluated these techniques on the nail hammering task (Figure~\ref{fig:camera_overlay}), comparing three models: one trained with both RC and VA, one with RC only, and one without either.
The combined use of RC and VA significantly expands the range of camera configurations under which the policy can successfully operate.

\noindent\textbf{Generalization:} \\
\noindent\uline{Spatial Generalization:} We evaluated spatial generalization by varying initial conditions across tasks: nail positions for hammering, meatball locations for scooping, goalkeeper setups for soccer ball kicking, and object poses across the pan for flipping (illustrated in Figure~\ref{fig:initial_overlay} in Appendix).

\noindent\uline{Object Generalization:}
Our method generalizes effectively to different objects in the pan-flipping task, including the egg and burger bun seen during training, and a 3D-printed meat patty (illustrated in Figure~\ref{fig:initial_overlay}, second column, in Appendix).
The policy learns to tilt the pan to slide the object into a corner, then flick it to achieve a successful flip, enabling robust generalization across object types.

\noindent\uline{Tool Generalization:} 
We evaluated tool generalization by testing the policy with five different pans: large, medium, small, tiny, and square.
The policy was trained using demonstrations with the large, medium, and square pans and evaluated on all five, with 12 trials per pan under varying initial configurations (illustrated in Figure~\ref{fig:diff_pan} in Appendix).
It achieved high success rates on the trained pans (large and medium). Performance declined on smaller pans due to limited surface area, and on the square pan due to shallow edges causing the bun to slide out during flipping.

\begin{figure*}[t] 
    \centering 
    \includegraphics[width=1\textwidth]{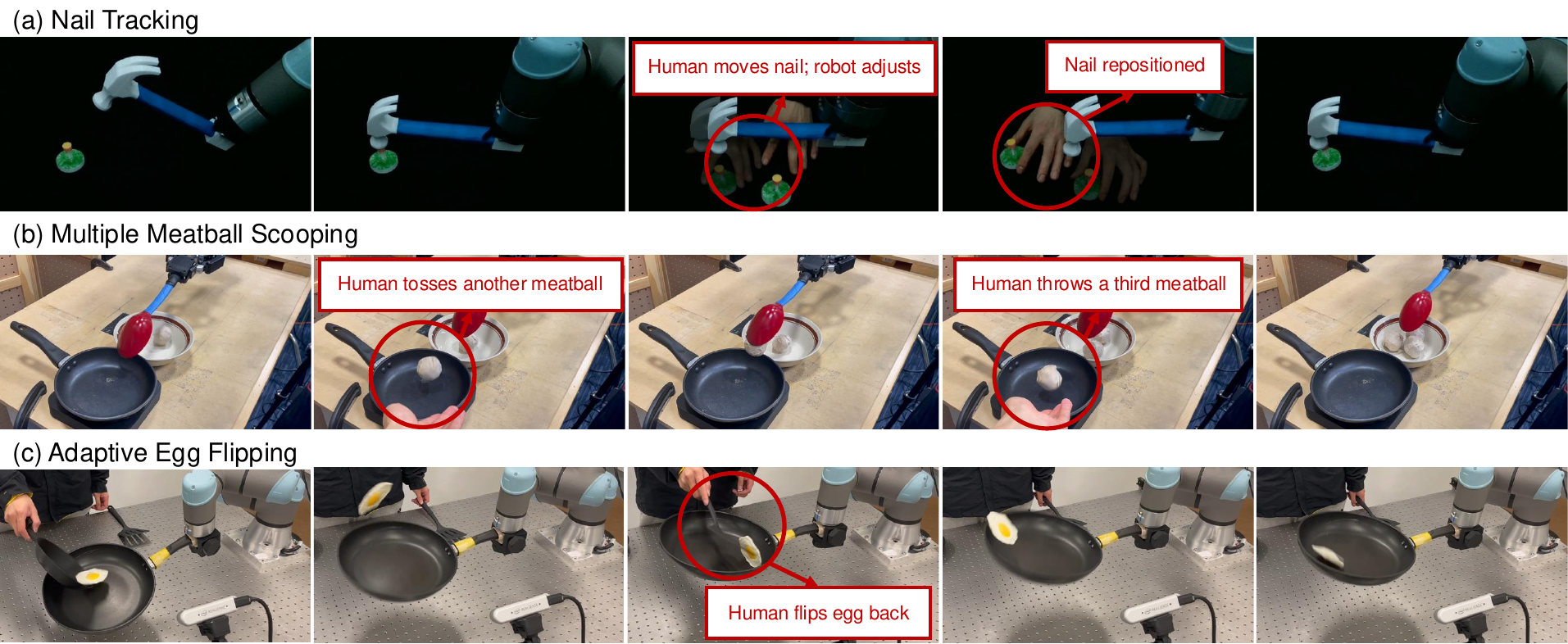} 
    \vspace{-0.23in} 
    \caption{\scriptsize{\textbf{Human Perturbation Robustness.}  
    The robot handles human-induced perturbations across three tasks: (1) In nail hammering, it tracked a manually moved nail; (2) In meatball scooping, it located and scooped new meatballs thrown mid-task; and (3) In egg flipping, it recovered the egg after human repositioning.}}
    \label{fig:human_perturb} 
    \vspace{-0.33in} 
\end{figure*}



\noindent\textbf{Robustness:}\\
\noindent\uline{Camera Pose Robustness:}
We evaluated the policy’s ability to handle camera pose variations by introducing camera shaking in three tasks: meatball scooping, nail hammering, and pan flipping (Figure~\ref{fig:robustness}(a)).
The first row shows the camera view, and the second shows the scene overview and shaking motion.
Despite disturbances, the policy consistently completed all tasks, enabled by random cropping during training, improving adaptation to partial views and minor visual changes.

\noindent\uline{Robot Base Robustness:}
To assess robustness to base movement, we manually shook the robot base during execution (Figure~\ref{fig:robustness}(b)).
When the shaking frequency exceeded the control frequency, the end effector oscillated with the base; however, the task-space action design enabled compensation and successful task completion.
As shown in Figure~\ref{fig:robustness}(d), the policy also maintained effectiveness under simultaneous camera and base shaking.

\noindent\uline{Chicken Head Stabilization:}
At lower shaking frequencies, where the perturbation was slower than the robot’s control loop, the end effector exhibited a stabilization behavior similar to a chicken’s head~\cite{XIA2025118649} (Figure~\ref{fig:robustness}(c)), maintaining steady control during mild base movements.

\noindent\uline{Human Perturbation Robustness:}
We evaluated resilience to human interventions (Figure~\ref{fig:human_perturb}).
The robot tracked moving nails, adapted to new meatballs thrown in mid-task, and re-flipped repositioned eggs, demonstrating robustness to real-time disturbances.

\begin{figure*}[t]
\centering
\includegraphics[width=1\textwidth]{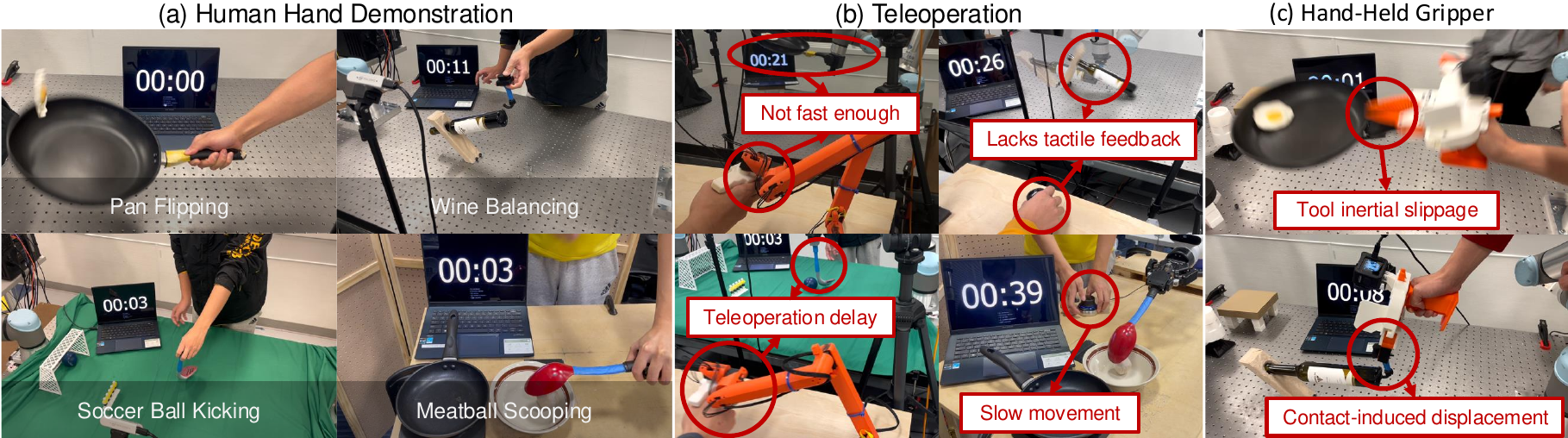}
\vspace{-0.23in}
\caption{\scriptsize{\textbf{Data Collection Efficiency and Reliability.}  
(a) Human hands excel in manipulation tasks, leveraging natural and intuitive efficiency.  
(b) Failure cases for Gello and Spacemouse include insufficient speed, lack of tactile feedback during data collection, safety stops, collisions, teleoperation delays, and difficulty handling high-speed or complex tasks.
(c) Failure cases for handheld grippers such as UMI~\cite{chi2024universal}, where issues arise from tool slippage due to inertia or displacement caused by contact forces.}}
\label{fig:data_collection}
\vspace{-0.25in}
\end{figure*}

\vspace{-0.12in}
\section{Data Collection Efficiency and Affordability}
\label{data_collection} 
\vspace{-0.1in}
We compare data collection methods for robot imitation learning across throughput, reliability, cost, usability, and precision. Full quantitative and qualitative analysis is provided in Appendix~\ref{appendix:data_collection}.

\begin{wrapfigure}{r}{0.48\textwidth}
\vspace{-2.5em}
\centering
\includegraphics[width=0.45\textwidth, trim=0pt 0pt 0pt 5pt, clip]{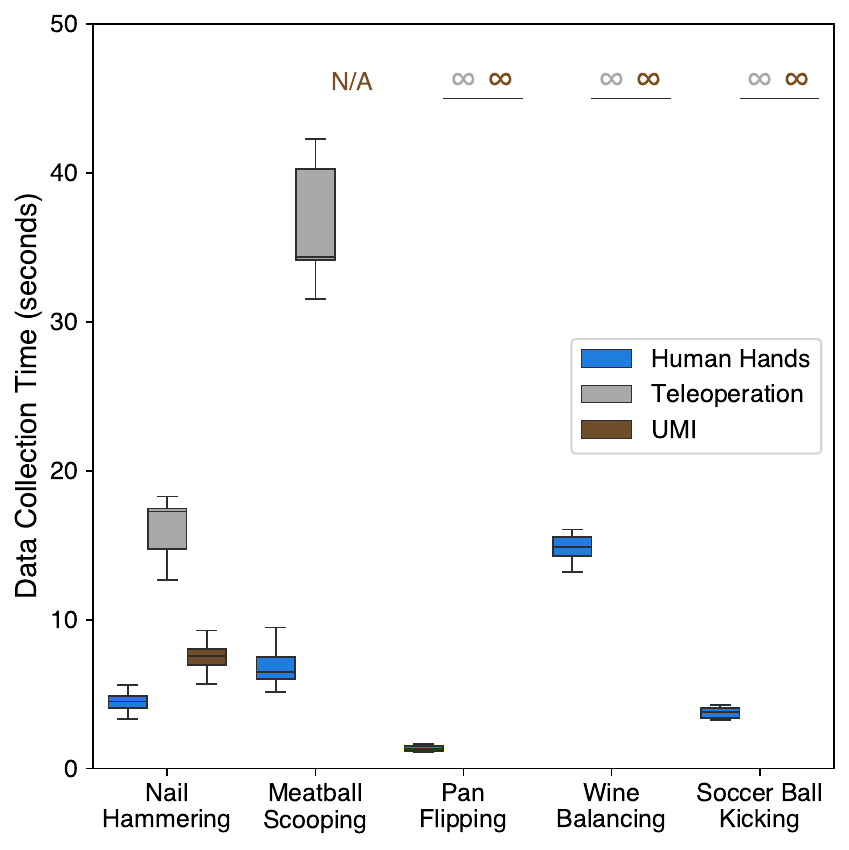}
\vspace{-0.7em}
\setlength{\belowcaptionskip}{-20pt}
\caption{\scriptsize\textbf{Quantitative Comparison of Data Collection Methods.}  
Human hands reduce data collection time by 73\% for nail hammering and 81\% for meatball scooping, while maintaining low variation.  
Teleoperation fails in dynamic and high-precision tasks.  
In nail hammering, human hands are 41\% faster than UMI~\cite{chi2024universal}, which also struggles with dynamic and low-texture environments.}
\setlength{\belowcaptionskip}{-10pt}
\label{fig:quali_data_collection}
\vspace{-1em}
\end{wrapfigure}

\noindent\textbf{Data Collection Efficiency:}
Leveraging the natural dexterity and intuitive control of human hands (Figure~\ref{fig:data_collection}(a)), our method achieves significantly higher data collection throughput compared to traditional methods. Quantitatively, human demonstrations reduce data collection time by \textbf{73\%} for nail hammering and \textbf{81\%} for meatball scooping relative to teleoperation methods (Gello, Spacemouse), as shown in Figure~\ref{fig:quali_data_collection}. Compared to handheld grippers (e.g., UMI~\cite{chi2024universal}), human demonstrations are \textbf{41\%} faster in nail hammering, and further succeed in dynamic tasks where both teleoperation and handheld methods consistently fail. This efficiency enables high-throughput, low-variance data collection critical for scalable robot learning.

\vspace{-0.05in}
\noindent\textbf{Reliability:} In contrast, Figures~\ref{fig:data_collection}(b) and (c) highlight typical failures for Gello, Spacemouse, and handheld grippers, including collisions, tool slippage, safety stops, and failures in dynamic or high-precision tasks. Teleoperation tools frequently suffer from latency, lack of tactile feedback, and difficulty handling rapid motions, leading to inconsistent trajectories and poor-quality demonstrations. Handheld grippers are prone to inertial slippage and loss of tool control under high forces, further limiting their applicability.
\par
\vspace{-0.05in}
\noindent\textbf{Cost and Usability:}
Our method requires no specialized hardware, incurring \textbf{zero additional cost} compared to teleoperation devices, handheld grippers, and VR/AR equipment, all of which demand substantial investment and maintenance. A detailed discussion is provided in Appendix~\ref{appendix:data_collection}, with a direct comparison summarized in Table~\ref{tab:data_cost}. The intuitive nature of human demonstrations further lowers the expertise barrier, making large-scale, high-quality data collection accessible without costly infrastructure.

By eliminating hardware dependence, minimizing operational failures, and dramatically improving throughput and precision, our human-centric data collection framework enables scalable, efficient, and reliable robot imitation learning.

\section{Conclusion} 
\label{sec:conclusion}
In this work, we presented a framework that enables robust policy training for diverse tool-use tasks by learning from human manipulation videos and bridging the embodiment gap they introduce.
Unlike traditional data collection methods, which are costly and hardware-dependent, our approach democratizes data collection by eliminating the need for specialized equipment or technical expertise, making large-scale robot learning accessible and scalable.
We validated the framework across challenging tasks, including nail hammering, meatball scooping, pan flipping, wine bottle balancing, and soccer ball kicking, demonstrating superior performance, robustness to variations in camera poses and base movements, and adaptability across 6-DOF and 7-DOF robots. By improving accessibility, scalability, and reliability, our work lays a strong foundation for advancing robotic manipulation in complex, real-world scenarios. \looseness=-1

\clearpage
\section{Limitations and Future Work} 
\label{sec:limitations}
Our framework has certain limitations. First, the perception pipeline relies on FoundationPose for extracting the tool's pose during manipulation. Errors in pose estimation may occasionally require data recollection, adding time and effort. Improving the reliability of the perception pipeline through more robust pose estimation algorithms or self-correction mechanisms is a promising direction for future work. 
Second, for novel view augmentation, significant noise and reduced realism are observed when augmented views deviate too far from the collected camera views, which can hinder policy performance. Future efforts could focus on leveraging advanced rendering techniques to enhance the realism of augmented views and improve policy generalization. 
Third, we assume the tool is rigidly attached to the robot's end effector; however, in real-world, contact-rich manipulation, minor shifts may occur, potentially affecting performance. Addressing this issue by incorporating tactile sensing could improve performance in contact-intensive tasks.
{Additionally, our method assumes a rigid tool and does not account for flexible or soft tools. Future work could explore using flexible representations for tool state estimation to better handle deformable tools in real-world manipulation scenarios.}


\acknowledgments{We thank Professor Wenzhen Yuan, Xiaoyu Zhang, Yucheng Mo, Yilong Niu, Hyoungju Lim, and Amin Mirzaee for their support and assistance in reproducing the UMI. We thank Professor Saurabh Gupta, Professor Junyi Geng, Lujie Yang, Neeloy Chakraborty, Hongkai Dai, Jacob Wagner, Shaoxiong Yao, Wei-Cheng Huang for their insightful feedback and suggestions.
This work was supported by ZJU-UIUC Joint Research Center Project  No. DREMES 202003, funded by Zhejiang University.
This research used the Delta advanced computing and data resource which is supported by the National Science Foundation (award OAC 2005572) and the State of Illinois. Delta is a joint effort of the University of Illinois Urbana-Champaign and its National Center for Supercomputing Applications.
Additionally, this work used NCSA Delta GPU at NCSA through allocation CIS240753 from the Advanced Cyberinfrastructure Coordination Ecosystem: Services \& Support (ACCESS) program, which is supported by U.S. National Science Foundation grants \#2138259, \#2138286, \#2138307, \#2137603, and \#2138296.
The robot experiments were carried out in part in the Center for Autonomy Robotics Laboratories at the University of Illinois Urbana-Champaign.
}


\bibliography{references}  

\begin{thebibliography}{68}
\providecommand{\natexlab}[1]{#1}
\providecommand{\url}[1]{\texttt{#1}}
\expandafter\ifx\csname urlstyle\endcsname\relax
  \providecommand{\doi}[1]{doi: #1}\else
  \providecommand{\doi}{doi: \begingroup \urlstyle{rm}\Url}\fi

\bibitem[Vaesen(2012)]{Vaesen_2012}
K.~Vaesen.
\newblock The cognitive bases of human tool use.
\newblock \emph{Behavioral and Brain Sciences}, 35\penalty0 (4):\penalty0 203–218, 2012.

\bibitem[Osiurak and Heinke(2018)]{OsiurakHeinke2018}
F.~Osiurak and D.~Heinke.
\newblock Looking for intoolligence: A unified framework for the cognitive study of human tool use and technology.
\newblock \emph{American Psychologist}, 73\penalty0 (2):\penalty0 169--185, 2018.

\bibitem[Bicchi and Kumar(2000)]{844081}
A.~Bicchi and V.~Kumar.
\newblock Robotic grasping and contact: a review.
\newblock In \emph{Proceedings 2000 ICRA. Millennium Conference. IEEE International Conference on Robotics and Automation. Symposia Proceedings (Cat. No.00CH37065)}, volume~1, pages 348--353 vol.1, 2000.

\bibitem[Prattichizzo and Trinkle(2016)]{Prattichizzo2016}
D.~Prattichizzo and J.~C. Trinkle.
\newblock \emph{Grasping}, pages 955--988.
\newblock Springer International Publishing, Cham, 2016.
\newblock ISBN 978-3-319-32552-1.

\bibitem[Carbone(2012)]{carbone2012grasping}
G.~Carbone.
\newblock \emph{Grasping in Robotics}.
\newblock Mechanisms and Machine Science. Springer London, 2012.
\newblock ISBN 9781447146643.

\bibitem[Lozano-Perez et~al.(1989)Lozano-Perez, Jones, Mazer, and O'Donnell]{16222}
T.~Lozano-Perez, J.~Jones, E.~Mazer, and P.~O'Donnell.
\newblock Task-level planning of pick-and-place robot motions.
\newblock \emph{Computer}, 22\penalty0 (3):\penalty0 21--29, 1989.

\bibitem[Lobbezoo et~al.(2021)Lobbezoo, Qian, and Kwon]{robotics10030105}
A.~Lobbezoo, Y.~Qian, and H.-J. Kwon.
\newblock Reinforcement learning for pick and place operations in robotics: A survey.
\newblock \emph{Robotics}, 10\penalty0 (3), 2021.
\newblock ISSN 2218-6581.

\bibitem[Fang et~al.(2019)Fang, Jia, Guo, Xu, Wen, and Sun]{Fang2019}
B.~Fang, S.~Jia, D.~Guo, M.~Xu, S.~Wen, and F.~Sun.
\newblock Survey of imitation learning for robotic manipulation.
\newblock \emph{International Journal of Intelligent Robotics and Applications}, 3\penalty0 (4):\penalty0 362--369, Dec 2019.
\newblock ISSN 2366-598X.

\bibitem[Hua et~al.(2021)Hua, Zeng, Li, and Ju]{s21041278}
J.~Hua, L.~Zeng, G.~Li, and Z.~Ju.
\newblock Learning for a robot: Deep reinforcement learning, imitation learning, transfer learning.
\newblock \emph{Sensors}, 21\penalty0 (4), 2021.
\newblock ISSN 1424-8220.

\bibitem[Jang et~al.(2022)Jang, Irpan, Khansari, Kappler, Ebert, Lynch, Levine, and Finn]{jang2022bc}
E.~Jang, A.~Irpan, M.~Khansari, D.~Kappler, F.~Ebert, C.~Lynch, S.~Levine, and C.~Finn.
\newblock Bc-z: Zero-shot task generalization with robotic imitation learning.
\newblock In \emph{Conference on Robot Learning (CoRL)}, volume 164, pages 991--1002. PMLR, 2022.

\bibitem[Wang et~al.(2023)Wang, Fan, Sun, Zhang, Fei-Fei, Xu, Zhu, and Anandkumar]{wang2023mimicplay}
C.~Wang, L.~Fan, J.~Sun, R.~Zhang, L.~Fei-Fei, D.~Xu, Y.~Zhu, and A.~Anandkumar.
\newblock Mimicplay: Long-horizon imitation learning by watching human play.
\newblock \emph{arXiv preprint arXiv:2302.12422}, 2023.

\bibitem[Brohan et~al.(2023)Brohan, Brown, Carbajal, Chebotar, Dabis, Finn, Gopalakrishnan, Hausman, Herzog, Hsu, et~al.]{brohan2022rt}
A.~Brohan, N.~Brown, J.~Carbajal, Y.~Chebotar, J.~Dabis, C.~Finn, K.~Gopalakrishnan, K.~Hausman, A.~Herzog, J.~Hsu, et~al.
\newblock Rt-1: Robotics transformer for real-world control at scale.
\newblock In \emph{Proceedings of Robotics: Science and Systems (RSS)}, 2023.

\bibitem[Seo et~al.(2024)Seo, Park, Yuan, Zhu, , and Sentis]{seo2024legato}
M.~Seo, H.~A. Park, S.~Yuan, Y.~Zhu, , and L.~Sentis.
\newblock Legato: Cross-embodiment visual imitation using a grasping tool, 2024.

\bibitem[Chi et~al.(2024)Chi, Xu, Pan, Cousineau, Burchfiel, Feng, Tedrake, and Song]{chi2024universal}
C.~Chi, Z.~Xu, C.~Pan, E.~Cousineau, B.~Burchfiel, S.~Feng, R.~Tedrake, and S.~Song.
\newblock Universal manipulation interface: In-the-wild robot teaching without in-the-wild robots.
\newblock In \emph{Proceedings of Robotics: Science and Systems (RSS)}, 2024.

\bibitem[Shao et~al.(2021)Shao, Migimatsu, Zhang, Yang, and Bohg]{shao2021concept2robot}
L.~Shao, T.~Migimatsu, Q.~Zhang, K.~Yang, and J.~Bohg.
\newblock Concept2robot: Learning manipulation concepts from instructions and human demonstrations.
\newblock \emph{The International Journal of Robotics Research}, 40\penalty0 (12-14):\penalty0 1419--1434, 2021.

\bibitem[Bahl et~al.(2022)Bahl, Gupta, and Pathak]{bahl2022human}
S.~Bahl, A.~Gupta, and D.~Pathak.
\newblock Human-to-robot imitation in the wild.
\newblock In \emph{Proceedings of Robotics: Science and Systems (RSS)}, 2022.

\bibitem[Pan et~al.(2023)Pan, Okorn, Zhang, Eisner, and Held]{pmlr-v205-pan23a}
C.~Pan, B.~Okorn, H.~Zhang, B.~Eisner, and D.~Held.
\newblock Tax-pose: Task-specific cross-pose estimation for robot manipulation.
\newblock In \emph{Proceedings of The 6th Conference on Robot Learning (CoRL)}, volume 205, pages 1783--1792. PMLR, 2023.

\bibitem[Connextion(2023)]{3dconnecxion2023spacemouse}
D.~Connextion.
\newblock Spacemouse, 2023.
\newblock URL \url{https://3dconnexion.com/us/spacemouse/}.

\bibitem[Wu et~al.(2023)Wu, Shentu, Lin, and Abbeel]{wu2023gello}
P.~Wu, F.~Shentu, X.~Lin, and P.~Abbeel.
\newblock {GELLO}: A general, low-cost, and intuitive teleoperation framework for robot manipulators.
\newblock In \emph{Towards Generalist Robots: Learning Paradigms for Scalable Skill Acquisition @ CoRL2023}, 2023.

\bibitem[Zhao et~al.(2020)Zhao, Queralta, and Westerlund]{Zhao2020SimtoRealTI}
W.~Zhao, J.~P. Queralta, and T.~Westerlund.
\newblock Sim-to-real transfer in deep reinforcement learning for robotics: a survey.
\newblock \emph{2020 IEEE Symposium Series on Computational Intelligence (SSCI)}, pages 737--744, 2020.

\bibitem[Peng et~al.(2018)Peng, Andrychowicz, Zaremba, and Abbeel]{8460528}
X.~B. Peng, M.~Andrychowicz, W.~Zaremba, and P.~Abbeel.
\newblock Sim-to-real transfer of robotic control with dynamics randomization.
\newblock In \emph{2018 IEEE International Conference on Robotics and Automation (ICRA)}, pages 3803--3810, 2018.

\bibitem[Ibarz et~al.(2021)Ibarz, Tan, Finn, Kalakrishnan, Pastor, and Levine]{doi:10.1177/0278364920987859}
J.~Ibarz, J.~Tan, C.~Finn, M.~Kalakrishnan, P.~Pastor, and S.~Levine.
\newblock How to train your robot with deep reinforcement learning: lessons we have learned.
\newblock \emph{The International Journal of Robotics Research}, 40\penalty0 (4-5):\penalty0 698--721, 2021.

\bibitem[Yu et~al.(2019)Yu, Kumar, Turk, and Liu]{8968053}
W.~Yu, V.~C. Kumar, G.~Turk, and C.~K. Liu.
\newblock Sim-to-real transfer for biped locomotion.
\newblock In \emph{2019 IEEE/RSJ International Conference on Intelligent Robots and Systems (IROS)}, pages 3503--3510, 2019.

\bibitem[Zhu et~al.(2022)Zhu, Joshi, Stone, and Zhu]{zhu2022viola}
Y.~Zhu, A.~Joshi, P.~Stone, and Y.~Zhu.
\newblock Viola: Imitation learning for vision-based manipulation with object proposal priors.
\newblock \emph{6th Annual Conference on Robot Learning (CoRL)}, 2022.

\bibitem[Chi et~al.(2023)Chi, Feng, Du, Xu, Cousineau, Burchfiel, and Song]{chi2023diffusionpolicy}
C.~Chi, S.~Feng, Y.~Du, Z.~Xu, E.~Cousineau, B.~Burchfiel, and S.~Song.
\newblock Diffusion policy: Visuomotor policy learning via action diffusion.
\newblock In \emph{Proceedings of Robotics: Science and Systems (RSS)}, 2023.

\bibitem[Chi et~al.(2024)Chi, Xu, Feng, Cousineau, Du, Burchfiel, Tedrake, and Song]{chi2024diffusionpolicy}
C.~Chi, Z.~Xu, S.~Feng, E.~Cousineau, Y.~Du, B.~Burchfiel, R.~Tedrake, and S.~Song.
\newblock Diffusion policy: Visuomotor policy learning via action diffusion.
\newblock \emph{The International Journal of Robotics Research}, 2024.

\bibitem[Seo et~al.(2023)Seo, Han, Sim, Bang, Gonzalez, Sentis, and Zhu]{seo2023trill}
M.~Seo, S.~Han, K.~Sim, S.~H. Bang, C.~Gonzalez, L.~Sentis, and Y.~Zhu.
\newblock Deep imitation learning for humanoid loco-manipulation through human teleoperation.
\newblock In \emph{IEEE-RAS International Conference on Humanoid Robots (Humanoids)}, 2023.

\bibitem[Wong et~al.(2021)Wong, Tung, Kurenkov, Mandlekar, Fei-Fei, and Savarese]{Wong2021ErrorAwareIL}
J.~Wong, A.~Tung, A.~Kurenkov, A.~Mandlekar, L.~Fei-Fei, and S.~Savarese.
\newblock Error-aware imitation learning from teleoperation data for mobile manipulation.
\newblock In \emph{Conference on Robot Learning}, 2021.

\bibitem[Zhao et~al.(2023)Zhao, Kumar, Levine, and Finn]{zhao2023learning}
T.~Z. Zhao, V.~Kumar, S.~Levine, and C.~Finn.
\newblock Learning fine-grained bimanual manipulation with low-cost hardware.
\newblock In \emph{Proceedings of Robotics: Science and Systems (RSS)}, 2023.

\bibitem[Fu et~al.(2024)Fu, Zhao, and Finn]{fu2024mobile}
Z.~Fu, T.~Z. Zhao, and C.~Finn.
\newblock Mobile aloha: Learning bimanual mobile manipulation with low-cost whole-body teleoperation.
\newblock \emph{arXiv preprint arXiv:2401.02117}, 2024.

\bibitem[Yu et~al.(2024)Yu, Han, Wang, Saxena, Xu, and Zhao]{yu2024mimictouch}
K.~Yu, Y.~Han, Q.~Wang, V.~Saxena, D.~Xu, and Y.~Zhao.
\newblock Mimictouch: Leveraging multi-modal human tactile demonstrations for contact-rich manipulation.
\newblock In \emph{8th Annual Conference on Robot Learning}, 2024.

\bibitem[Bahety et~al.(2024)Bahety, Mandikal, Abbatematteo, and Mart{\'\i}n-Mart{\'\i}n]{bahety2024screwmimic}
A.~Bahety, P.~Mandikal, B.~Abbatematteo, and R.~Mart{\'\i}n-Mart{\'\i}n.
\newblock Screwmimic: Bimanual imitation from human videos with screw space projection.
\newblock In \emph{Robotics: Science and Systems (RSS), 2024}, 2024.

\bibitem[Wen et~al.(2022)Wen, Lian, Bekris, and Schaal]{wen2022yodo}
B.~Wen, W.~Lian, K.~Bekris, and S.~Schaal.
\newblock You only demonstrate once: Category-level manipulation from single visual demonstration.
\newblock In \emph{Proceedings of Robotics: Science and Systems (RSS)}, 2022.

\bibitem[Hu et~al.(2022)Hu, Huang, Rybkin, and Jayaraman]{hu2022know}
E.~S. Hu, K.~Huang, O.~Rybkin, and D.~Jayaraman.
\newblock Know thyself: Transferable visual control policies through robot-awareness.
\newblock In \emph{International Conference on Learning Representations (ICLR)}, 2022.

\bibitem[Salhotra et~al.(2023)Salhotra, xI~Chun Arthur~Liu, and Sukhatme]{salhotra2023bridging}
G.~Salhotra, xI~Chun Arthur~Liu, and G.~Sukhatme.
\newblock Bridging action space mismatch in learning from demonstrations.
\newblock \emph{arXiv preprint arXiv:2304.03833}, 2023.

\bibitem[Yang et~al.(2023)Yang, Sadigh, and Finn]{yang2023polybot}
J.~H. Yang, D.~Sadigh, and C.~Finn.
\newblock Polybot: Training one policy across robots while embracing variability.
\newblock In \emph{Annual Conference on Robot Learning (CoRL)}, 2023.

\bibitem[Devin et~al.()Devin, Gupta, Darrell, Abbeel, and Levine]{devin2016learning}
C.~Devin, A.~Gupta, T.~Darrell, P.~Abbeel, and S.~Levine.
\newblock Learning modular neural network policies for multi-task and multi-robot transfer.
\newblock In \emph{International Conference on Robotics and Automation (ICRA)}.

\bibitem[Wang et~al.(2018)Wang, Liao, and Jimmy~Ba]{wang2016nervenet}
T.~Wang, R.~Liao, and S.~F. Jimmy~Ba.
\newblock Nervenet: Learning structured policy with graph neural networks.
\newblock In \emph{International Conference on Learning Representations (ICLR)}, 2018.

\bibitem[Yu et~al.(2017)Yu, Tan, Liu, and Turk]{yu2017preparing}
W.~Yu, J.~Tan, C.~K. Liu, and G.~Turk.
\newblock Preparing for the unknown: Learning a universal policy with online system identification.
\newblock In \emph{Robotics: Science and Systems (RSS)}, 2017.

\bibitem[Chen et~al.(2019)Chen, Murali, and Gupta]{chen2019hardware}
T.~Chen, A.~Murali, and A.~Gupta.
\newblock Hardware conditioned policies for multi-robot transfer learning, 2019.

\bibitem[Ghadirzadeh et~al.(2021)Ghadirzadeh, Chen, Poklukar, Finn, Björkman, and Kragic]{ghadirzadeh2021bayesian}
A.~Ghadirzadeh, X.~Chen, P.~Poklukar, C.~Finn, M.~Björkman, and D.~Kragic.
\newblock Bayesian meta-learning for few-shot policy adaptation across robotic platforms.
\newblock In \emph{International Conference on Intelligent Robots and Systems (IROS)}, 2021.

\bibitem[Attarian et~al.(2023)Attarian, Asif, Liu, Hari, Garg, Gilitschenski, and Tompson]{attarian2023geometry}
M.~Attarian, M.~A. Asif, J.~Liu, R.~Hari, A.~Garg, I.~Gilitschenski, and J.~Tompson.
\newblock Geometry matching for multi-embodiment grasping, 2023.

\bibitem[Sferrazza et~al.(2024)Sferrazza, Huang, Liu, Lee, and Abbeel]{sferrazza2024body}
C.~Sferrazza, D.-M. Huang, F.~Liu, J.~Lee, and P.~Abbeel.
\newblock Body transformer: Leveraging robot embodiment for policy learning.
\newblock In \emph{8th Annual Conference on Robot Learning}, 2024.

\bibitem[Yang et~al.(2024)Yang, Glossop, Bhorkar, Shah, Vuong, Finn, Sadigh, and Levine]{yang2024pushinglimitscrossembodimentlearning}
J.~Yang, C.~Glossop, A.~Bhorkar, D.~Shah, Q.~Vuong, C.~Finn, D.~Sadigh, and S.~Levine.
\newblock Pushing the limits of cross-embodiment learning for manipulation and navigation, 2024.

\bibitem[Wen et~al.(2024)Wen, Lin, So, Chen, Dou, Gao, and Abbeel]{wen2023anypoint}
C.~Wen, X.~Lin, J.~So, K.~Chen, Q.~Dou, Y.~Gao, and P.~Abbeel.
\newblock Any-point trajectory modeling for policy learning.
\newblock In \emph{Proceedings of Robotics: Science and Systems (RSS)}, 2024.

\bibitem[Lynch et~al.(2020)Lynch, Khansari, Xiao, Kumar, Tompson, Levine, and Sermanet]{lynch2020learning}
C.~Lynch, M.~Khansari, T.~Xiao, V.~Kumar, J.~Tompson, S.~Levine, and P.~Sermanet.
\newblock Learning latent plans from play.
\newblock In \emph{Conference on robot learning}, pages 1113--1132. PMLR, 2020.

\bibitem[Kareer et~al.(2024)Kareer, Patel, Punamiya, Mathur, Cheng, Wang, Hoffman, and Xu]{kareer2024egomimic}
S.~Kareer, D.~Patel, R.~Punamiya, P.~Mathur, S.~Cheng, C.~Wang, J.~Hoffman, and D.~Xu.
\newblock Egomimic: Scaling imitation learning via egocentric video.
\newblock \emph{ArXiv}, abs/2410.24221, 2024.

\bibitem[Sermanet et~al.(2017)Sermanet, Lynch, Hsu, and Levine]{sermanet2018time}
P.~Sermanet, C.~Lynch, J.~Hsu, and S.~Levine.
\newblock Time-contrastive networks: Self-supervised learning from multi-view observation.
\newblock In \emph{2017 IEEE Conference on Computer Vision and Pattern Recognition Workshops (CVPRW)}, pages 486--487, 2017.

\bibitem[Chen et~al.(2021)Chen, Abbeel, and Pathak]{chen2021unsupervised}
B.~Chen, P.~Abbeel, and D.~Pathak.
\newblock Unsupervised learning of visual 3d keypoints for control.
\newblock In \emph{ICML}, 2021.

\bibitem[Shridhar et~al.(2022)Shridhar, Manuelli, and Fox]{shridhar2023perceiver}
M.~Shridhar, L.~Manuelli, and D.~Fox.
\newblock Perceiver-actor: A multi-task transformer for robotic manipulation.
\newblock In \emph{Proceedings of the 6th Conference on Robot Learning (CoRL)}, 2022.

\bibitem[Gervet et~al.(2023)Gervet, Xian, Gkanatsios, and Fragkiadaki]{gervet2023act3d}
T.~Gervet, Z.~Xian, N.~Gkanatsios, and K.~Fragkiadaki.
\newblock Act3d: 3d feature field transformers for multi-task robotic manipulation.
\newblock In \emph{7th Annual Conference on Robot Learning}, 2023.

\bibitem[Zhu et~al.(2023)Zhu, Jiang, Stone, and Zhu]{zhu2023learning}
Y.~Zhu, Z.~Jiang, P.~Stone, and Y.~Zhu.
\newblock Learning generalizable manipulation policies with object-centric 3d representations.
\newblock \emph{ArXiv}, abs/2310.14386, 2023.

\bibitem[Sadeghi et~al.(2018)Sadeghi, Toshev, Jang, and Levine]{sadeghi2018sim2realservo}
F.~Sadeghi, A.~Toshev, E.~Jang, and S.~Levine.
\newblock Sim2real viewpoint invariant visual servoing by recurrent control.
\newblock In \emph{2018 IEEE/CVF Conference on Computer Vision and Pattern Recognition}, pages 4691--4699, 2018.

\bibitem[Zhou et~al.(2023)Zhou, Kim, Wang, Florence, and Finn]{zhou2023nerf}
A.~Zhou, M.~J. Kim, L.~Wang, P.~Florence, and C.~Finn.
\newblock { NeRF in the Palm of Your Hand: Corrective Augmentation for Robotics via Novel-View Synthesis }.
\newblock In \emph{2023 IEEE/CVF Conference on Computer Vision and Pattern Recognition (CVPR)}, pages 17907--17917, Los Alamitos, CA, USA, June 2023. IEEE Computer Society.

\bibitem[Sharma et~al.(2019)Sharma, Pathak, and Gupta]{NEURIPS2019_8a146f1a}
P.~Sharma, D.~Pathak, and A.~Gupta.
\newblock Third-person visual imitation learning via decoupled hierarchical controller.
\newblock In H.~Wallach, H.~Larochelle, A.~Beygelzimer, F.~d\textquotesingle Alch\'{e}-Buc, E.~Fox, and R.~Garnett, editors, \emph{Advances in Neural Information Processing Systems}, volume~32. Curran Associates, Inc., 2019.

\bibitem[Chen et~al.(2024)Chen, Xu, Dharmarajan, Irshad, Cheng, Keutzer, Tomizuka, Vuong, and Goldberg]{chen2024roviaug}
L.~Y. Chen, C.~Xu, K.~Dharmarajan, M.~Z. Irshad, R.~Cheng, K.~Keutzer, M.~Tomizuka, Q.~Vuong, and K.~Goldberg.
\newblock Rovi-aug: Robot and viewpoint augmentation for cross-embodiment robot learning.
\newblock In \emph{Conference on Robot Learning (CoRL)}, Munich, Germany, 2024.

\bibitem[Kirillov et~al.(2023)Kirillov, Mintun, Ravi, Mao, Rolland, Gustafson, Xiao, Whitehead, Berg, Lo, Doll{\'a}r, and Girshick]{kirillov2023segany}
A.~Kirillov, E.~Mintun, N.~Ravi, H.~Mao, C.~Rolland, L.~Gustafson, T.~Xiao, S.~Whitehead, A.~C. Berg, W.-Y. Lo, P.~Doll{\'a}r, and R.~Girshick.
\newblock Segment anything.
\newblock \emph{arXiv:2304.02643}, 2023.

\bibitem[Zhang et~al.(2023)Zhang, Rao, and Agrawala]{zhang2023adding}
L.~Zhang, A.~Rao, and M.~Agrawala.
\newblock Adding conditional control to text-to-image diffusion models, 2023.

\bibitem[Sargent et~al.(2023)Sargent, Li, Shah, Herrmann, Yu, Zhang, Chan, Lagun, Fei-Fei, Sun, and Wu]{zeronvs}
K.~Sargent, Z.~Li, T.~Shah, C.~Herrmann, H.-X. Yu, Y.~Zhang, E.~R. Chan, D.~Lagun, L.~Fei-Fei, D.~Sun, and J.~Wu.
\newblock {ZeroNVS}: Zero-shot 360-degree view synthesis from a single real image.
\newblock \emph{CVPR, 2024}, 2023.

\bibitem[Yuan et~al.(2024)Yuan, Wei, Cheng, Zhang, Chen, and Xu]{yuan2024learning}
Z.~Yuan, T.~Wei, S.~Cheng, G.~Zhang, Y.~Chen, and H.~Xu.
\newblock Learning to manipulate anywhere: A visual generalizable framework for reinforcement learning.
\newblock In \emph{8th Annual Conference on Robot Learning}, 2024.

\bibitem[Leroy et~al.(2024)Leroy, Cabon, and Revaud]{mast3r_arxiv24}
V.~Leroy, Y.~Cabon, and J.~Revaud.
\newblock Grounding image matching in 3d with mast3r, 2024.

\bibitem[Wen et~al.(2024)Wen, Yang, Kautz, and Birchfield]{foundationposewen2024}
B.~Wen, W.~Yang, J.~Kautz, and S.~Birchfield.
\newblock {FoundationPose}: Unified 6d pose estimation and tracking of novel objects.
\newblock In \emph{CVPR}, 2024.

\bibitem[Young et~al.(2020)Young, Gandhi, Tulsiani, Gupta, Abbeel, and Pinto]{young2020visual}
S.~Young, D.~Gandhi, S.~Tulsiani, A.~Gupta, P.~Abbeel, and L.~Pinto.
\newblock Visual imitation made easy, 2020.

\bibitem[Ren et~al.(2024)Ren, Liu, Zeng, Lin, Li, Cao, Chen, Huang, Chen, Yan, Zeng, Zhang, Li, Yang, Li, Jiang, and Zhang]{ren2024grounded}
T.~Ren, S.~Liu, A.~Zeng, J.~Lin, K.~Li, H.~Cao, J.~Chen, X.~Huang, Y.~Chen, F.~Yan, Z.~Zeng, H.~Zhang, F.~Li, J.~Yang, H.~Li, Q.~Jiang, and L.~Zhang.
\newblock Grounded sam: Assembling open-world models for diverse visual tasks, 2024.

\bibitem[Jonsson(2024)]{Jonsson2024}
P.~Jonsson.
\newblock Ericsson mobility report november 2024.
\newblock In \emph{Ericsson Mobility Report}. Ericsson, 2024.

\bibitem[Xia et~al.(2025)Xia, Li, Wen, Xu, Wang, and Zhang]{XIA2025118649}
S.~Xia, Y.~Li, G.~Wen, D.~Xu, K.~Wang, and H.~Zhang.
\newblock Natural mechanism of superexcellent vibration isolation of the chicken neck.
\newblock \emph{Journal of Sound and Vibration}, 594:\penalty0 118649, 2025.
\newblock ISSN 0022-460X.

\bibitem[Meta(2023)]{quest}
Meta.
\newblock Meta quest, 2023.
\newblock URL \url{https://www.meta.com/quest}.

\bibitem[Apple(2024)]{vision_pro}
Apple.
\newblock Apple vision pro, 2024.
\newblock URL \url{https://www.apple.com/apple-vision-pro/}.

\end{thebibliography}

\clearpage

\newpage
\appendix
\begin{center}
    {\LARGE \bfseries APPENDIX}
\end{center}
\vspace{1em}
\tableofcontents

\newpage

\section{Design Choice}
\subsection{Key Capabilities and Practical Benefits}

Our framework enables the direct transfer of human manipulation data into deployable robot policies. It is designed to fulfill the following key objectives:

\begin{itemize}

    \item \textbf{Support for Dynamic and High-Precision Tasks:} 
    Human manipulation, with its inherent fluidity, enables the execution of highly dynamic tasks. Examples include flipping an egg in a pan or performing other actions that require swift, accurate, and natural motions — challenges that are often difficult to address with traditional teleoperation systems or handheld grippers.

    \item \textbf{Robustness:} 
    The framework ensures robust performance under dynamic conditions, enabling reliable task execution even with moving or shaking cameras. While broader deployment on mobile platforms such as quadrupeds or humanoids remains an open challenge, our design and experimental results suggest strong potential for generalization to dynamic environments.

    \item \textbf{Generalization Across Robotic Embodiments and Object Categories:} 
    The framework demonstrates broad generalizability, validated on robotic platforms such as the UR5e and Kinova Gen3. It extends its capabilities to manipulate a wide range of object categories, showcasing its adaptability to various tasks, setups, and environments.

    \item \textbf{Affordability and Accessibility:} 
    The framework requires only two monocular RGB cameras, such as smartphones, webcams, or RealSense cameras. With approximately 7.14 billion smartphones worldwide — covering around 90\% of the global population — this setup is accessible to almost anyone~\cite{Jonsson2024}. By relying solely on RGB cameras, the framework eliminates the need for designing, printing, or manufacturing additional hardware during the data collection, ensuring a cost-effective and inclusive solution.

    \item \textbf{Intuitive and Natural Interaction:} 
    Users can interact naturally, without the need for specialized equipment or additional tools. Using their bare hands and common tools, participants can intuitively perform a variety of tasks. Our approach removes technical barriers associated with 3D printing and other hardware setups, fostering a seamless, user-friendly experience for data collection.

\end{itemize}

\section{Detailed Experiment Setup}
\subsection{Task Descriptions}
\label{appendix:tasks}
\noindent\textbf{Nail Hammering:}
The task involves hammering a 3D-printed nail, requiring the robot to locate the nail, draw back the hammer, and strike the nail tip accurately. With a diameter of less than 15.5 mm, the nail tip demands high precision. 
Challenges include localizing the nail tip precisely and planning effective hammer trajectories. To evaluate generalization, the initial position of the nail is varied across different spatial configurations. We collected 180 seconds of data (40 episodes) from a single participant.

\noindent\textbf{Meatball Scooping:}
In this task, the robot must use a spoon to scoop a meatball from a pan and transfer it to a bowl. This task is challenging due to the complex dynamics of the meatball, which can roll unpredictably within the pan. Additionally, the interaction between the spoon and the meatball requires careful control, as improper contact can cause the meatball to slip or escape the spoon. We randomize the initial position of the meatball within the pan to test its generalization capability. We collected 340 seconds of data (50 episodes) from a single participant.

\noindent\textbf{Pan Flipping (Egg, Burger Bun, Meat Patty):}
The objective of this task is to use a pan to flip various objects, such as an egg, a burger bun, and a meat patty. The task is challenging due to its high-speed dynamics, requiring the robot to overcome gravity and accurately manage the interaction between the pan and the objects. Each object differs in weight, shape, and texture, adding further complexity.
This task evaluates the policy’s ability to handle fast, contact-rich interactions and adapt to diverse object types. To increase variability, the initial positions of the objects within the pan are randomized. Furthermore, the rapid and dynamic nature of the task makes it unsuitable for classical demonstration collection methods, highlighting the advantages of using bare-handed human videos for data collection. We collected 50 seconds of data (38 episodes) from a single participant using three different pans and two object types.

\noindent\textbf{Wine Balancing:}
In this task, the robot needs to use a hook to lift a wine bottle and carefully insert it into an unstable, zero-gravity wine rack. The task is challenging due to the precise control required to suspend the bottle in mid-air and counteract gravitational forces effectively. Any over-insertion or under-insertion will cause the bottle to lose balance. To constrain the horizontal movement of the rack, screws were added as obstacles to limit lateral motion. No additional variability was introduced. We collected 223 seconds of data (15 episodes) from a single participant.

\noindent\textbf{Soccer Ball Kicking:}
In this task, the robot must use a golf club to kick a ball that slides into a field and direct it into the goal. To increase the challenge, a 3D-printed row of players serves as obstacles between the robot and the goal. The task is difficult because the robot must accurately intercept the moving ball, strike it with the correct force and direction, and ensure it avoids obstacles before reaching the goal. The position of the player obstacle varies. We collected 78 seconds of data (20 episodes) from a single participant.
\begin{table*}[tb]
\centering
\vspace{-0.1in}
\caption{\scriptsize{\textbf{Benchmark Attributes of Real-World Tasks.} These benchmarks evaluate the precision, adaptability, and capability of our framework to address tasks requiring high precision, handling extreme dynamics, utilizing extrinsic dexterity, performing in contact-rich scenarios, and overcoming gravity.}}
\label{table:benchmark_attributes}
\vspace{-0.05in}
\resizebox{\textwidth}{!}{%
\begin{tabular}{l|ccccc}
\toprule
Benchmark & High-Precision & Extreme Dynamics & Using Extrinsic Dexterity & Contact-Rich & Overcoming Gravity \\
\midrule
Task 1: Nail Hammering & \textcolor{ggreen}{\CheckmarkBold} & \textcolor{gred}{\CustomDash} & \CustomDash & \CustomDash & \CustomDash \\
Task 2: Meatball Scooping & \textcolor{ggreen}{\CheckmarkBold} & \CustomDash & \textcolor{ggreen}{\CheckmarkBold} & \textcolor{ggreen}{\CheckmarkBold} & \CustomDash \\
Task 3: Pan Flipping (Egg, Bun, Patty) & \CustomDash & \textcolor{ggreen}{\CheckmarkBold} & \textcolor{ggreen}{\CheckmarkBold} & \textcolor{ggreen}{\CheckmarkBold} & \textcolor{ggreen}{\CheckmarkBold} \\
Task 4: Wine Balancing & \textcolor{ggreen}{\CheckmarkBold} & \CustomDash & \textcolor{ggreen}{\CheckmarkBold} & \textcolor{ggreen}{\CheckmarkBold} & \textcolor{ggreen}{\CheckmarkBold} \\
Task 5: Soccer Ball Kicking & \CustomDash & \CustomDash & \CustomDash & \textcolor{ggreen}{\CheckmarkBold} & \CustomDash \\
\bottomrule
\end{tabular}}
\vspace{-0.11in}
\end{table*}

\begin{figure*}[tb] 
    \centering
    \includegraphics[width=1\textwidth]{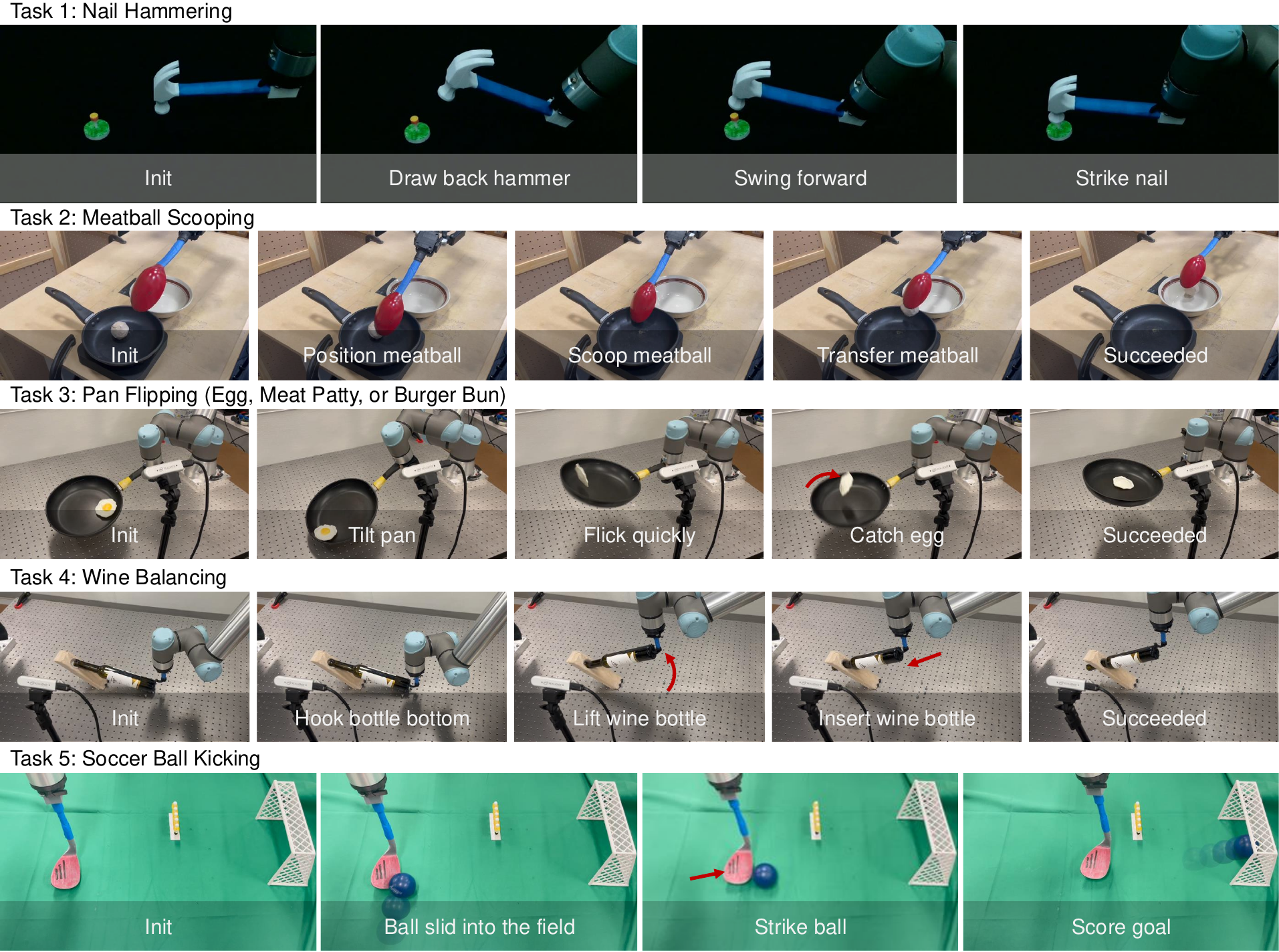}
    \vspace{-0.22in}
    \caption{\scriptsize{\textbf{Policy Rollouts.} We evaluate diverse real-world tasks: nail hammering (precision in locating a nail tip), meatball scooping (slippery object, constrained environments), pan flipping (extremely dynamic, high-speed, contact-rich), wine balancing (precise control of unstable objects), and soccer ball kicking (dynamic object handling, goal-directed actions).}}
    \label{fig:policy_rollout}
    \vspace{-0.28in}
\end{figure*}

\begin{figure}[tb]
    \centering
    \includegraphics[width=0.4\textwidth]{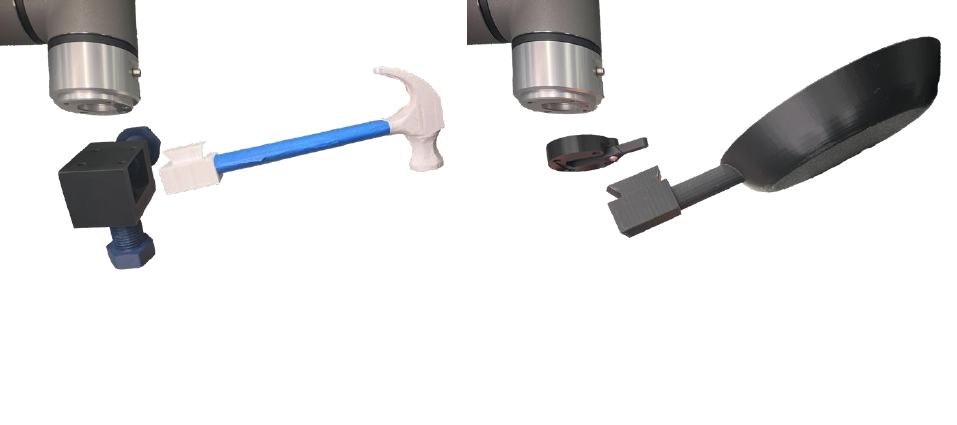}
    \vspace{-0.2in}
    \caption{\scriptsize{\textbf{Fast Tool Changer.} Two designs are shown: the left accommodates general tools with a screw mechanism, and the right clips onto tools with specific mounting shapes.}}
    \label{fig:tool_changer}
    \vspace{-0.05in}
\end{figure}

\begin{figure}[tb] 
    \centering
    \includegraphics[width=0.50\textwidth]{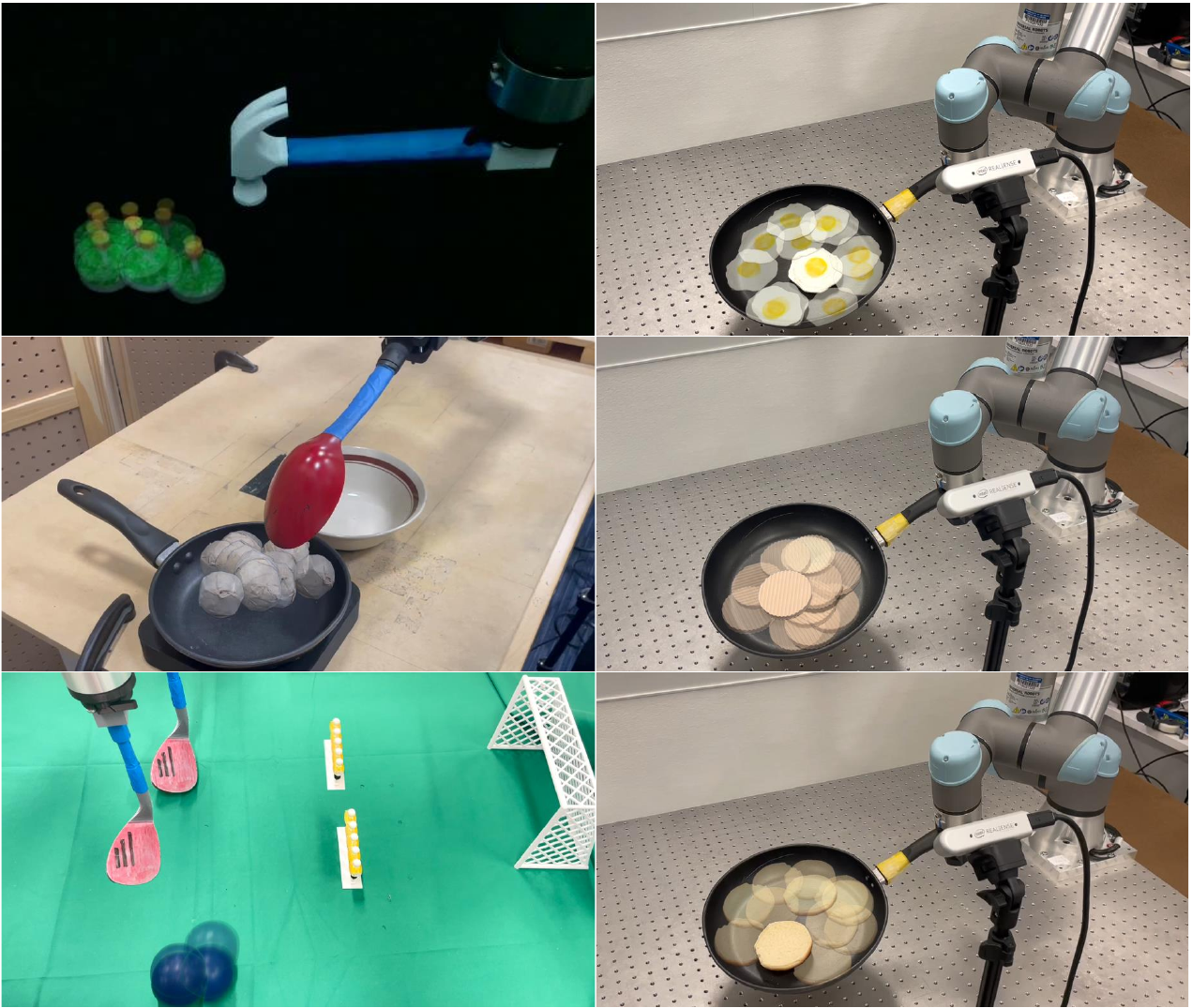}
    \vspace{-0.5em}
    \caption{\scriptsize\textbf{Initial States for All Evaluation Episodes.} All methods are evaluated using the same set of manually defined initial states, overlaid in the image. These states ensure diverse variations to test the policy’s spatial generalization capabilities.}
    \label{fig:initial_overlay}
    \vspace{-1em}
\end{figure}

\subsection{Implementation Details}
\subsubsection{Hardware Design}
\label{appendix:hardware_design}
We designed two fast tool changers compatible with robots using the ISO 9409-1-50-4-M6 flange, as shown in Figure~\ref{fig:tool_changer}. The left design utilizes a screw mechanism to accommodate general tools, while the right design employs clips for tools with specific mounting shapes.

\subsubsection{Tool Pose Estimation}
We use Polycam to scan the tool and obtain its mesh. The mesh is later feed into FoundationPose~\cite{foundationposewen2024} for 6D pose estimation.

\begin{figure}[t]
\centering
\begin{minipage}{0.48\textwidth}
    \centering
    \includegraphics[width=\textwidth]{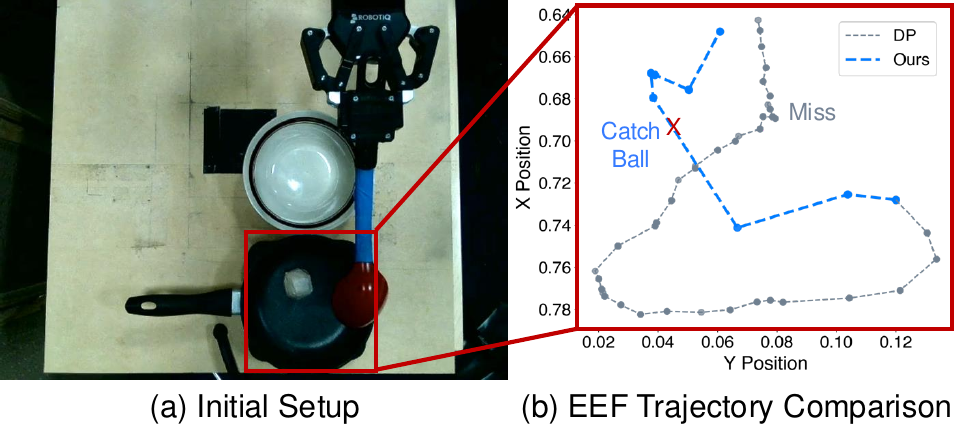}
    \caption{\scriptsize\textbf{Policy Execution Trajectory Comparison.} (a) Initial setup for meatball scooping. (b) Comparison of end-effector XY trajectories from our framework and a policy trained on robot-collected data.}
    \label{fig:trajectory_comparison}
\end{minipage}
\hfill
\begin{minipage}{0.48\textwidth}
    \centering
    \includegraphics[width=\textwidth]{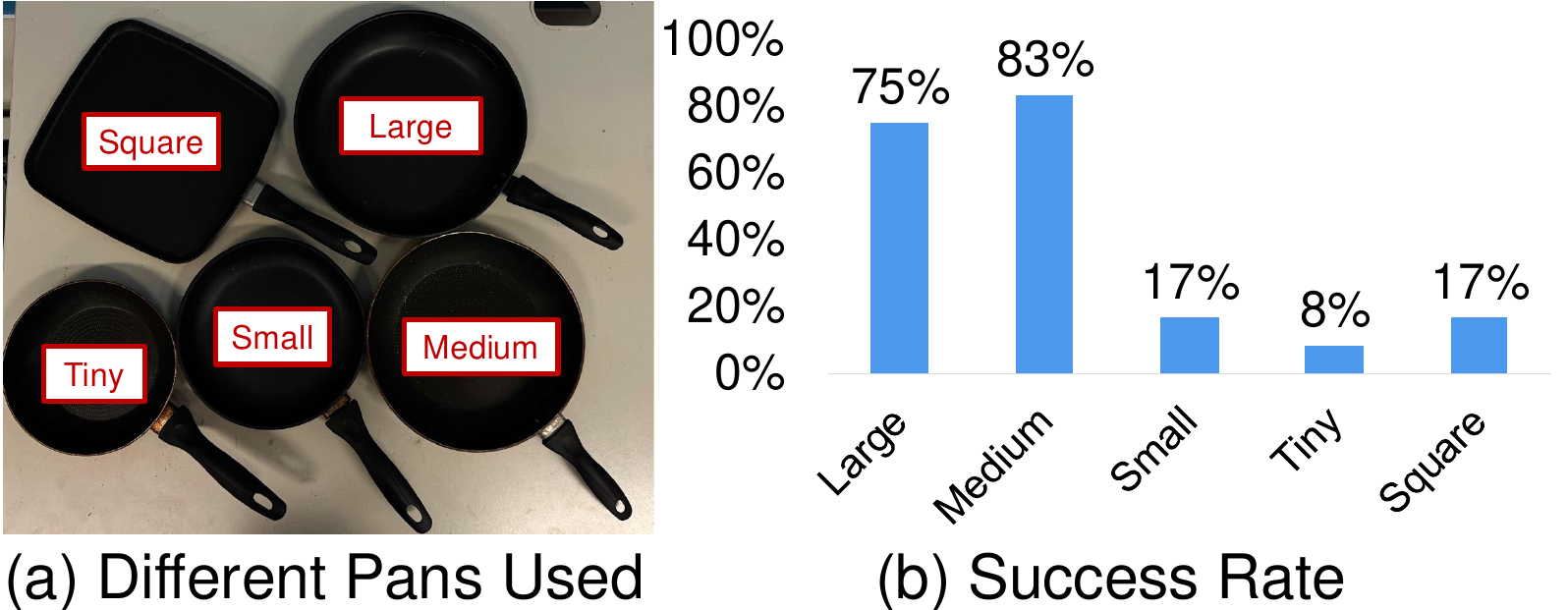}
    \caption{\scriptsize\textbf{Tool Generalization.} (a) The tested pans. (b) Success rate across 12 testing trials.}
    \label{fig:diff_pan}
\end{minipage}
\end{figure}

\begin{figure}[t]
    \centering
    \includegraphics[width=0.5\textwidth]{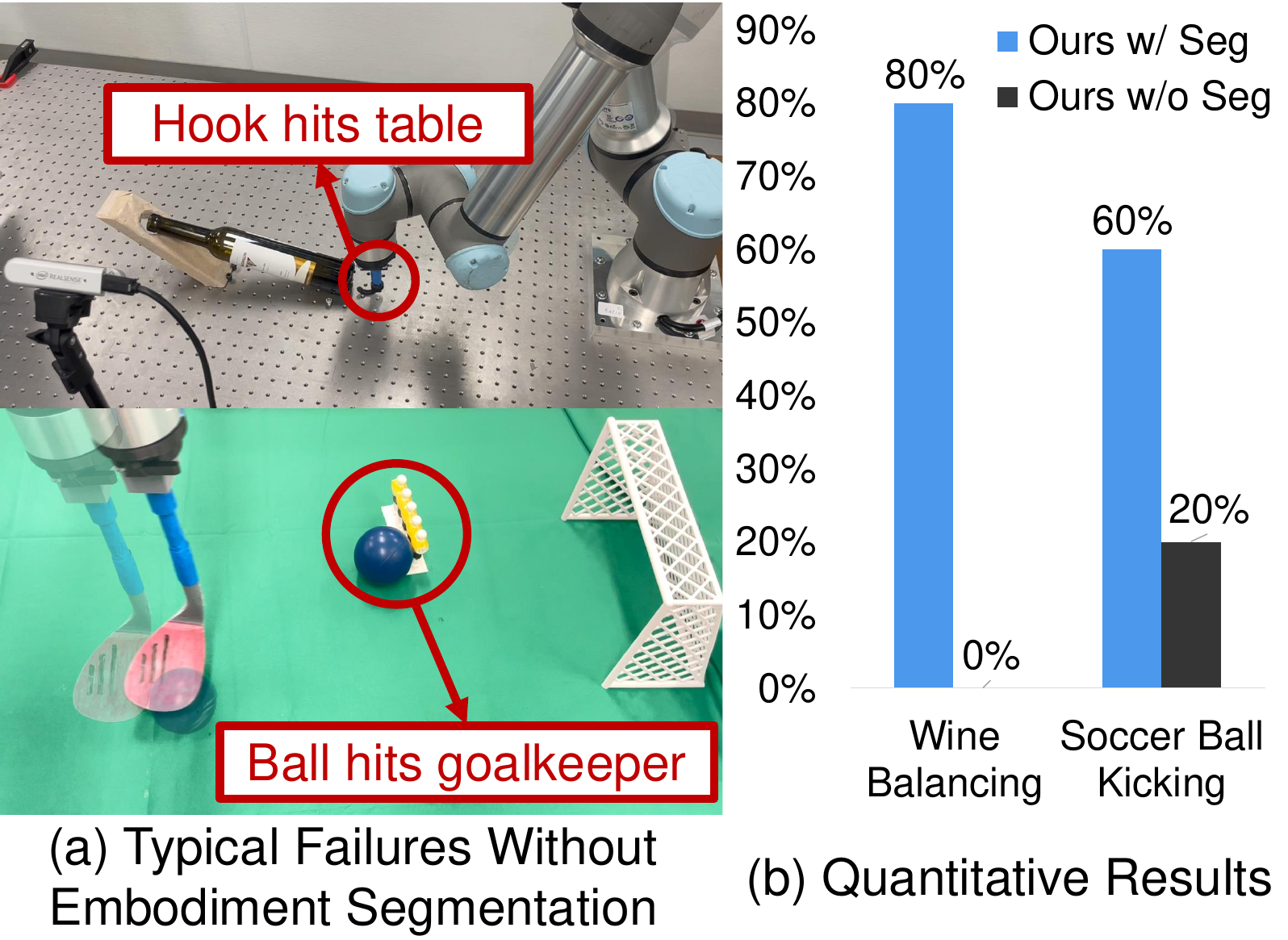}
    \vspace{1\baselineskip}
    \vspace{-0.2in}
    \caption{\scriptsize{\textbf{Effects of Embodiment Segmentation.} (a) Failure cases without segmentation: In the wine balancing task, the robot strikes the table, triggering safety stops. In the soccer ball kicking task, it performs shorter, less precise actions. (b) Quantitative results: Segmentation improved success rates in wine balancing (8 vs. 0) and soccer ball kicking (6 vs. 2) by reducing the visual gap between training and deployment.}}
    \label{fig:wo_seg_failure}
    \vspace{-0.2in}
\end{figure}

\section{Additional Experimental Results}
\label{appendix:additional_experimental}

\noindent\textbf{Policy Execution Trajectory Comparison:}\label{appendix:execution_smooth} 
Our framework produces faster, smoother, and more natural trajectories compared to traditional approaches, as shown by the end-effector (EEF) XY trajectory for the meatball scooping task in Figure~\ref{fig:trajectory_comparison}.
Figure~\ref{fig:trajectory_comparison}(a) shows the task setup, and Figure~\ref{fig:trajectory_comparison}(b) compares our policy rollout with a baseline trained on robot-collected data.
Our trajectory is significantly smoother, with 10$\times$ fewer waypoints, resulting in more fluid execution, reduced cumulative errors, and improved sample efficiency, thereby mitigating the distribution shifts commonly observed in behavior cloning. In contrast, the baseline exhibits excessive waypoints and discontinuous motions that hinder precise task execution.

\noindent\textbf{Effects of Embodiment Segmentation:}
Embodiment Segmentation masks the agent's embodiments during data collection and policy deployment, ensuring visually consistent scenes and reducing the training-deployment visual gap. Embodiment Segmentation significantly improves policy performance, as shown in Figure~\ref{fig:wo_seg_failure}.  
Figure~\ref{fig:wo_seg_failure}(a) highlights failure cases without segmentation. In the wine balancing task, the robot strikes the table, triggering safety stops due to improper bottle handling. In the soccer ball kicking task, the robot's actions are inconsistent, shorter, and less precise than during training. Quantitative results in Figure~\ref{fig:wo_seg_failure}(b) further underscore segmentation's impact. Across 10 trials, segmentation enabled 8 successes in the wine balancing task, while the model without it achieved none. Similarly, in the soccer ball kicking task, segmentation resulted in 6 successes, compared to 2 without it. By aligning training and testing visual distributions, Embodiment Segmentation ensures consistent and reliable robot performance during the training and deployment.

\section{Detailed Analysis on Data Collection Efficiency and Affordability}
\label{appendix:data_collection} 

We compare various data collection methods for robot imitation learning, focusing on throughput, reliability, cost, usability, and precision. Our evaluation includes teleoperation tools like Gello and Spacemouse for 6DOF (UR5e) and 7DOF (Kinova Gen3) robots, alongside methods such as Visual Imitation Made Easy, handheld grippers (e.g., UMI and LEGATO), and devices like VR (Meta Quest 2), AR (Apple Vision Pro), and Kinematic replicate (Gello).

\subsection{Data Collection Efficiency}  
Our framework achieves significantly higher data collection throughput than traditional methods, enabling more demonstrations within the same timeframe. The improvement is driven by the natural and intuitive efficiency of human manipulation, which ensures faster and more reliable task execution. Figure~\ref{fig:data_collection}(a) highlights the superior manipulation capabilities of human hands, while Figure~\ref{fig:quali_data_collection} quantifies the substantial time savings per episode.  
For nail hammering and meatball scooping, Gello and Spacemouse were used as teleoperation methods, respectively. Human hands reduced data collection time by 73\% and 81\% for nail hammering and meatball scooping, with consistently low variation in performance. In more complex tasks like pan flipping, wine balancing, and soccer ball kicking, teleoperation methods failed entirely due to limitations such as lack of tactile feedback, delays, and difficulty handling dynamic or precise actions.  Our method further reduces data collection time by 41\% compared to handheld grippers such as UMI~\cite{chi2024universal} in nail hammering. UMI proved ineffective in wine balancing and pan flipping due to tool inertial slippage or contact-induced displacement, and failed in soccer kicking because of difficulty localizing large, fast motions. Moreover, it requires rich textures to build a pre-collection map, which our method does not. These results underscore the superior efficiency, robustness, and versatility of human manipulation as a scalable solution for high-quality robot learning datasets.

\subsection{Reliability}
Figure~\ref{fig:data_collection}(b) and Figure~\ref{fig:data_collection}(c) illustrates typical failure cases with Gello, Spacemouse, and UMI~\cite{chi2024universal}, which frequently encounter issues such as safety stops or collisions during data collection. In contrast, our method ensures smooth, uninterrupted operation, avoiding these limitations. Traditional methods face significant challenges in high-speed or complex tasks. For example, Gello and Spacemouse struggle with replicating the extreme dynamics and precise motions required for flipping objects like eggs during pan flipping, often resulting in unsuccessful attempts. Similarly, teleoperation delays prevent timely strikes during soccer ball kicking, consistently leading to missed kicks and repeated failures. In tasks like wine balancing, the absence of tactile feedback impairs precision during the data collection, causing the wine bottle to tip over during data collection. Furthermore, in meatball scooping, the velocity vectors generated by Spacemouse input lead to jerky trajectories with redundant waypoints, significantly reducing efficiency. These challenges make effective training impractical with traditional methods. By leveraging human manipulation, our framework not only addresses these limitations but also provides a reliable and scalable solution for dynamic and precision-demanding tasks.

\subsection{Discussion of Data Collection Methods}
\label{appendx:dis_data_collection}
Table~\ref{tab:data_cost} compares various data collection methods based on cost, usability, expertise requirements, intuitiveness, and precision. Our method incurs no additional cost (\$0), unlike hardware-dependent solutions like UMI and LEGATO, which demand significant investment. This affordability makes our approach accessible to users from diverse backgrounds without financial constraints.
Unlike hardware-based systems such as UMI, LEGATO, Gello, and Spacemouse, which are prone to malfunctions and maintenance issues, our hardware-free framework ensures reliability and eliminates repair delays or expenses. Additionally, it requires no supplementary 3D printing, in contrast to approaches like Visual Imitation Made Easy, UMI, and LEGATO. The simplicity of our design promotes inclusivity in collecting large-scale dataset for robot learning research.
Our method also offers a more natural experience compared to tools like Spacemouse, while being far more cost-effective than VR and AR devices. Moreover, systems like Gello and Spacemouse lack the precision necessary for dynamic tasks, a limitation addressed by our approach.
Overall, our method is a cost-effective, and accessible solution for data collection, overcoming key drawbacks of existing approaches while reducing complexity and maintenance needs.

\begin{table*}[tb]
\centering
\vspace{-0.1in}
\caption{\scriptsize{\textbf{Comparison of Data Collection Methods.} This table compares various data collection methods for robotics. For cost, we calculate only the additional expenses required for data collection, excluding cameras, as they are considered a basic and commonly used sensor for robots rather than an additional purchase. Each method is assessed based on cost, ease of use, required expertise, precision, and maintenance effort. Our method stands out as cost-free, easy to use, highly precise, and requiring minimal maintenance.}}
\label{tab:data_cost}
\vspace{-0.05in}
\resizebox{\textwidth}{!}{%
\begin{tabular}{l|ccccc}
\toprule
\textbf{Method} & \textbf{Cost} & \textbf{Ready-to-Use} & \textbf{Pre-Knowledge Required} & \textbf{Precise} & \textbf{Maintenance Expense} \\
\midrule
Visual Imitation Made Easy~\cite{young2020visual} & \$340  & No  & Yes & No  & Moderate \\
UMI~\cite{chi2024universal}                        & \$371  & No  & Yes & \textbf{Yes} & Moderate \\
LEGATO~\cite{seo2024legato}                        & \$1060 & No  & Yes & \textbf{Yes} & Moderate \\
Spacemouse~\cite{3dconnecxion2023spacemouse}       & \$169  & \textbf{Yes} & Yes & \textbf{Yes} & Low \\
VR (Meta Quest 2~\cite{quest})                     & \$300  & \textbf{Yes} & Yes & No  & Moderate \\
AR (Apple Vision Pro~\cite{vision_pro})            & \$3499 & \textbf{Yes} & Yes & \textbf{Yes} & High \\
Gello~\cite{wu2023gello}                           & \$272  & No  & Yes & No  & Moderate \\
Ours                                               & \textbf{\$0} & \textbf{Yes} & \textbf{No} & \textbf{Yes} & \textbf{Minimal} \\
\bottomrule
\end{tabular}}
\vspace{-0.1in}
\end{table*}

\end{document}